\documentclass{article}
\usepackage[a4paper, total={17cm, 25cm}]{geometry}

\usepackage{graphicx}  
\usepackage{authblk}  
\usepackage{xcolor}
\usepackage{soul}  
\sethlcolor{white}  

\usepackage{biblatex}  
\usepackage{amsmath}  
\usepackage{hyperref}
\usepackage{booktabs}  

\usepackage{pifont}
\newcommand{\cmark}{\ding{51}}  

\usepackage{array}  
\usepackage{multirow}

\usepackage{adjustbox}  

\usepackage{float}  

\title{Deep learning-based compression of giga-resolution whole slide images}
\author[1,2]{Maren Høibø}
\author[3]{Etienne Gaucher}
\author[3,4]{Ingerid Reinertsen}
\author[1,2]{Marit Valla}
\author[3,4]{Erik Smistad}

\affil[1]{Department of Clinical and Molecular Medicine, Norwegian University of Science and Technology (NTNU), NO-7491 Trondheim, Norway}
\affil[2]{Clinic of Laboratory Medicine, St. Olavs hospital, Trondheim University Hospital, NO-7030 Trondheim, Norway}
\affil[3]{Department of Health Research, SINTEF Digital, NO-7465 Trondheim, Norway}
\affil[4]{Department of Circulation and Medical Imaging, Norwegian University of Science and Technology (NTNU), NO-7491 Trondheim, Norway}

\usepackage{fancyhdr}
\begin{document}
\pagestyle{fancy}
\fancyhf{}
\fancyhead[L]{This paper is currently under consideration in a scientific journal.}
\date{}  

\maketitle
\thispagestyle{fancy}

\section*{Abstract}
Implementation of digital pathology leads to an increased number of whole slide images (WSIs). The large size of WSIs is challenging. Today, WSIs are compressed with codecs like JPEG resulting in several gigabytes per WSI, and large amounts of space \hl{are} wasted\hl{ }storing glass. 
In this study, deep learning-based tissue segmentation for glass removal, and deep learning\hl{ }compression methods were explored and compared with JPEG, JPEG-2000 and JPEG-XL\hl{.}
Image pyramids (N=21) with intact glass,\hl{ }glass replaced by single-colored pixels, and\hl{ }glass replaced by zero-byte tiles\hl{ }were created and compressed with JPEG, JPEG-XL and a deep learning model. Additionally, several compression models were evaluated on a tissue patch dataset and compared with JPEG, JPEG-2000 and JPEG-XL.
Removing glass reduced file sizes considerably for JPEG and JPEG-XL. 
Deep learning-based \hl{image} compression \hl{reduced the WSI size by 43-72\% compared to JPEG compression, whereas deep learning-based glass removal reduced the WSI size by 0.3-33\%, and 6-62\% using only single-colored pixels and removing all-glass tiles, respectively. Combining the two gave a small improvement to a 44-80\% total size reduction which indicates that deep learning-based image compression is able to efficiently compress glass tiles, whereas JPEG is not. }
On the tissue patch dataset, the best deep learning-based compression models saved on average $\sim$35-40\% per patch compared to JPEG, while keeping an average SSIM above 0.95, whereas JPEG-XL and JPEG-2000 saved 17\% and 14\%, respectively while keeping an SSIM of 0.96.
However, the deep learning models had higher decompression times than JPEG and JPEG-XL.\\
\\
\noindent\textbf{Keywords:} Autoencoder, Deep Learning, Image Compression, JPEG, JPEG-2000, JPEG-XL, Pathology, Tissue Segmentation, Whole Slide Image.

\section{Introduction}
\label{sec:introduction}
In diagnostic pathology, tissue slides are assessed by pathologists, and the diagnostic pathology report often guides treatment of patients. Scanned tissue sections, whole slide images (WSIs), enable the use of digital tools and artificial intelligence (AI)-based analysis of tissue slides~\cite{Salto-TellezManuel2019Aitr}. A challenge in digital pathology is the large size of the images. 
In order to efficiently view and process WSIs, they are usually stored in a \textit{tiled image pyramid} data structure, where several downsampled versions of the WSI are stored, and each level is divided into a large set of equally sized image tiles.
Tissue slides scanned at $\times$40 magnification can have image sizes of up to $200,000\times100,000$ pixels \cite{10.3389/fmed.2022.971873}. Without compression, this equals a size of more than 55 GB for a single image. 

In addition, a biopsy often includes several sections, and a combination of hematoxylin and eosin (HE)-stained and immunohistochemistry sections are often used to determine the diagnosis. Thus, one biopsy can result in many WSI files. Without efficient compression of the WSIs, storage capacity will be a huge challenge for pathology laboratories in the future. Reducing the size of WSIs could help reduce digital storage space, energy consumption, and costs at hospitals. It could also decrease transfer time of images\hl{ }and reduce training time of machine learning models by enabling more images to fit on solid-state drives (SSD). SSDs are considerably faster, but with smaller storage capacity than conventional hard drives.

With \textit{lossless} compression, in which no image detail is lost or changed, the file size of WSIs would be too large for practical use. 
The solution is to use \textit{lossy} compression, and accept some loss in image detail. Although it was introduced in 1992, JPEG 
still remains one of the most popular lossy compression methods\hl{ }and is used in digital pathology. JPEG uses the discrete cosine transform with Huffman coding~\cite{SANTACRUZ2002113}, and can compress images with a factor of about 10, thus reducing the total WSI size from over 50 GB to 5 GB with little noticeable degradation in image quality. 
The JPEG algorithm has a quality parameter ranging from 0 to 100, which controls the trade-off between image quality and file size reduction. A higher value results in better image quality, but larger compressed size. 
In 2000, a new JPEG format, \textit{JPEG-2000}, was released. It uses wavelet transformation and arithmetic coding~\cite{952804}, and promised better compression performance~\cite{jpeg200Rabbani2009}. 
More recently, \textit{JPEG-XL}~\cite{alakuijala2019jpeg} was introduced as a replacement for JPEG, claiming even better compression rates.
However, to the best of our knowledge, this new compression algorithm has not been used commercially in the medical imaging domain. 

In recent years, the success of deep learning within image processing has prompted several research groups to investigate the use of deep learning and convolutional neural networks (CNNs) to learn efficient image compression with techniques such as autoencoders. One can hypothesize that a CNN can represent a highly efficient domain-specific WSI compression model when trained on pathology data. A domain-specific model may be more efficient than general purpose models trained on arbitrary images.

For natural images, deep learning-based compression has shown promise compared to JPEG~\cite{BalléJohannes2017EOIC, BalléJohannes2018Vicw, JohnstonNick2018ILIC,RippelOren2017RAIC, TheisLucas2017LICw,MinnenDavid2018JAaH, mentzer2020highfidelitygenerativeimagecompression,HeDailan2022EELI}. Autoencoders and generative adversarial network (GAN)-based models have achieved good results when evaluated with image quality metrics~\cite{BalléJohannes2018Vicw,mentzer2020highfidelitygenerativeimagecompression}. Ballé \textit{et al.} presented a factorized and a hyperprior compression model based on autoencoder architectures~\cite{BalléJohannes2017EOIC, BalléJohannes2018Vicw}. The factorized model consisted of a convolutional autoencoder with a lossy quantization and an arithmetic encoder and decoder.  
The hyperprior model extended the factorized model with the addition of a scale hyperprior as side information through an additional autoencoder~\cite{BalléJohannes2018Vicw}. Minnen \textit{et al.} further built on these models with an additional context model (autoregressive), including the mean value from the hyperprior when combining the hyperprior with the context model~\cite{MinnenDavid2018JAaH}. 

In histopathology, Barsi \textit{et al.} created a compression model based on an autoencoder combined with a classifier~\cite{BarsiAgnes2024Adlc}, whereas Fisher \textit{et al.}\cite{fischer23} implemented the factorized compression model by Ballé \textit{et al.}~\cite{BalléJohannes2018ENTf} with a custom feature similarity in their loss function using the CompressAI framework~\cite{begaint2020compressai}. Fischer \textit{et al.} trained the model on 25 000 histopathological patches from WSIs from breast and colon. Their model outperformed JPEG with quality parameter 80, with a 76\% file size reduction, while keeping the multi scale structural similarity index (MS-SSIM) at 0.99~\cite{fischer23}. Fischer \textit{et al.}~\cite{FischerMaximilian2024LICf} trained a model consisting of a stain encoder/decoder, encoding both the RGB and stain channels before they were compressed with a compression autoencoder. In a downstream task (classification of patches as tumor or non-tumor) they outperformed JPEG at high compression~\cite{FischerMaximilian2024LICf}.

\begin{figure}[h!]
    \centering
    \includegraphics[width=0.5\linewidth]{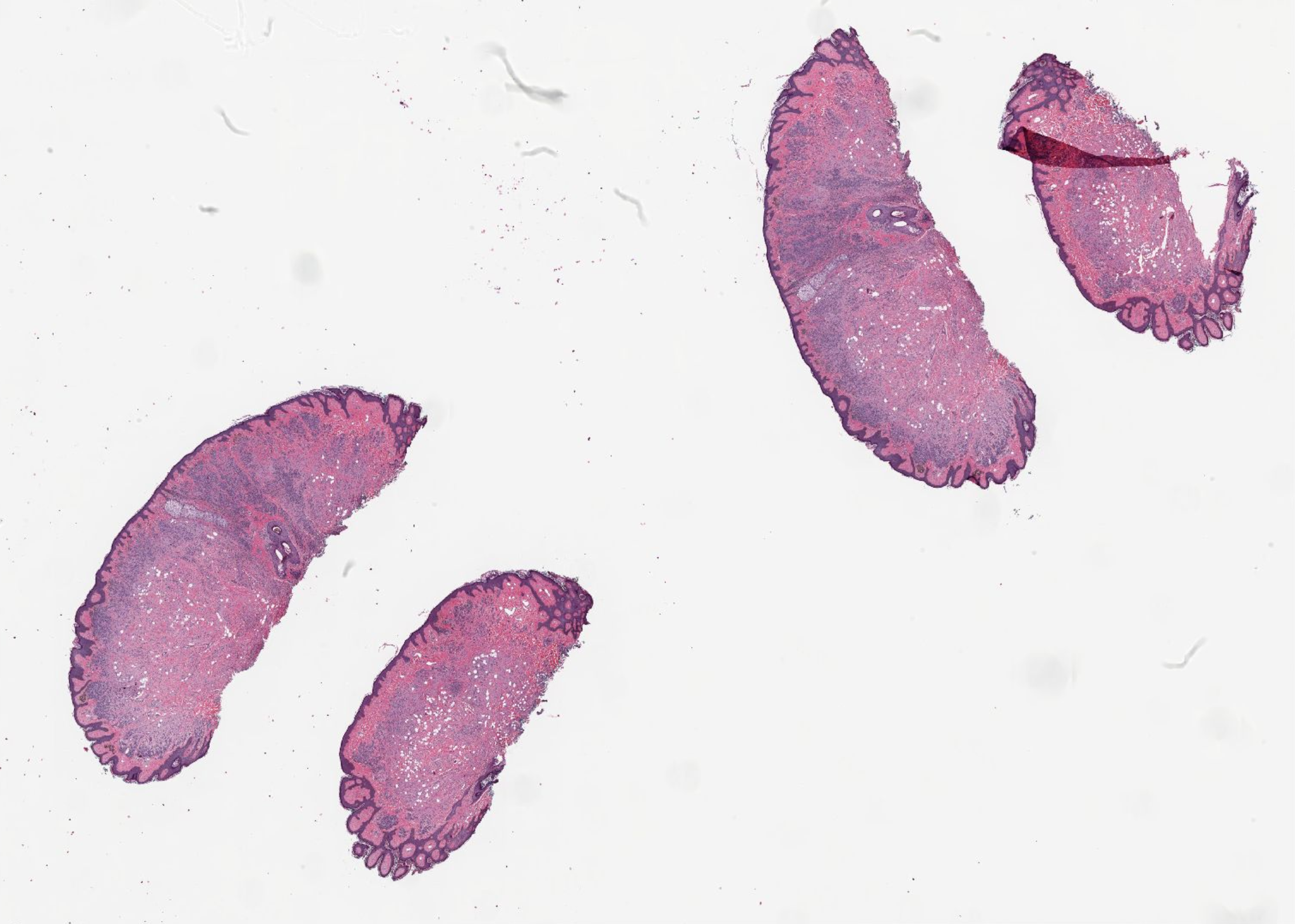}
    \caption{Histopathology images may include large areas of glass, here shown in gray, potentially wasting several gigabytes in storing image pixels that do not include any diagnostic information.
    The method proposed in this paper uses deep learning to segment tissue. A new WSI where the glass is removed is then created.
    Image \textit{CMU-1} from the OpenSlide Aperio SVS Test Data~\cite{aperioTestDataOpenSlide} (\href{https://creativecommons.org/publicdomain/zero/1.0/}{CC0} license.)}
    \label{fig:cmu-1}
\end{figure}

Whereas deep learning-based compression shows potential, few have attempted to make solutions for full-scale WSI compression. In a full-scale WSI, many pixels may not contain tissue, and thus hold no diagnostic information (Fig. \ref{fig:cmu-1}). Compressing irrelevant background areas more than tissue areas has been tried with JPEG-2000~\cite{helin2018optimized}. However, even highly compressed, these glass-only areas are a waste of storage space and should ideally be identified and fully removed from the image pyramid, to optimize storage space. This is a task suited for CNNs, which excel at image segmentation tasks such as finding tissue and glass in images. The resulting tissue areas can then be compressed using deep learning.

The contributions of this paper are:

\begin{itemize}
    \item A comprehensive evaluation of file size reduction and image quality of WSI tissue patches using different compression codecs (JPEG, JPEG-2000, JPEG-XL) and AI compression models.
    \item A comprehensive evaluation of WSI compression.
    \item Open software for reading and writing tiled pyramid WSIs, segmenting tissue, and reading and writing empty (e.g. glass) tiles using the TIFF file format. The software implementation is available in the FAST framework for both Python and C++~\cite{SmistadErik2015Fffh,SmistadErik2019HPNN}.
    \item An open AI model for accurate segmentation of tissue in HE-stained WSIs trained on open data. 
    \item The training and evaluation code, tissue segmentation model, and a Python script for easy WSI conversion and removal of glass are available at \\
    \url{https://github.com/AICAN-Research/whole-slide-image-compression}.
\end{itemize}

\section{Methods}

\begin{figure*}[h!]
    \centering
    \includegraphics[width=0.9\linewidth]{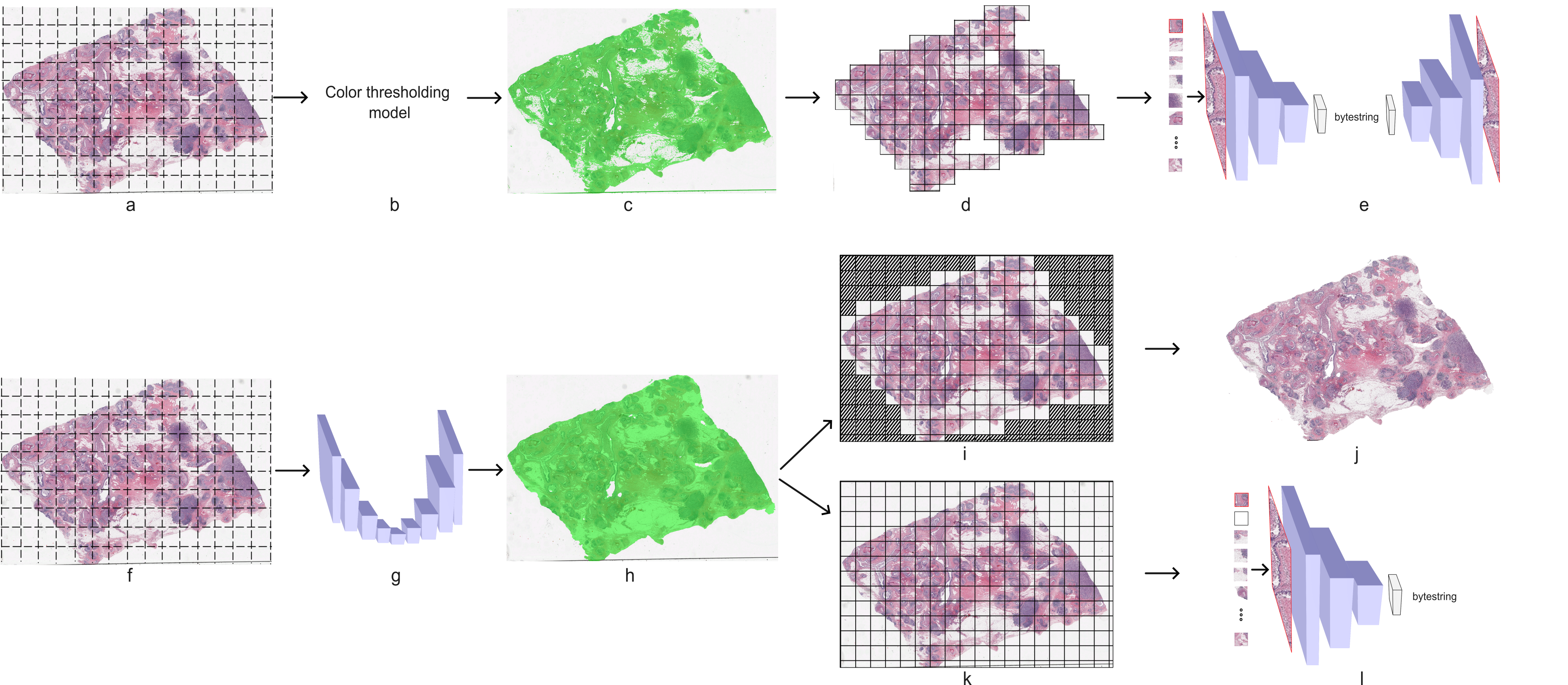}
    \caption{Illustration of the proposed method. Top row: \textit{Tissue compression:} A color thresholding segmentation model was used (a-c) to generate patches from within the tissue areas only (d). The patches were used to train and evaluate deep learning models. JPEG, JPEG-2000, and JPEG-XL compression were also evaluated on the patches. Image quality metrics and size reduction were evaluated. 
    Bottom row: \textit{Tissue segmentation and glass removal (f-j)} By inferring a patch-wise U-Net-based tissue segmentation model (f-g), tissue areas were identified in the WSI (h). New tiled image pyramids were created with the same pyramid structure as the original WSIs. Based on the segmentation, all tiles including only glass were identified (i) and replaced by empty tiles or single colored tiles in the new JPEG or JPEG-XL compressed image pyramid (j). \textit{Glass removal and tissue compression (f-h, k-l)} The segmentation model was inferred on the original WSI (h), and patches with size 256$\times$256 pixels were extracted (k). The patches were compressed to bytestrings through a neural network (purple) and quantization and entropy coder (light gray) (l). The bytestrings were saved on disk in \textit{patch pyramid} folders, to compare size reduction of these \textit{patch pyramids} with the JPEG-compressed image pyramids. The neural network-based tissue compression was only explored on patches, and support for neural network-based tissue compression in WSIs was not implemented in this study. The WSI in this illustration is the Her2Pos\_Case\_37 from the HER2 tumor ROIs dataset~\cite{Farahmand22} available at The Cancer Imaging Archive (TCIA)\cite{clark2013cancer}, with a \href{https://creativecommons.org/licenses/by/4.0/}{CC BY 4.0} license, and has been modified for this illustration. 
    }
    \label{fig:overview_proposedMethod}
\end{figure*}

In this study, we investigated two techniques for reducing the file size of WSIs while keeping a high image quality.
The first was to use deep learning to segment tissue in the WSIs, followed by removal of the identified glass regions. The second was to use modern lossy compression techniques such as deep learning and JPEG-XL to compress tissue areas. \hl{Fig.} \ref{fig:overview_proposedMethod} shows the process of applying these two methods to an existing WSI. 

\subsection{Data}\label{sec:dataset}
A dataset of 21 HE-stained WSIs from seven different organs was established (Supplementary material Table 1)~\cite{Farahmand22,CPTAC_coad_20,CMB-CRC22,CMB-LCA22,CPTAC-OV20,CPTAC-PDA18,CMB-PCA22,CPTAC-UCEC19}. All WSIs were from The Cancer Imaging Archive (TCIA) \cite{clark2013cancer} and were publicly available. The slides were scanned at magnification $\times 20$ or $\times 40$. The majority of the WSIs were JPEG-compressed, whereas a few were compressed with JPEG-2000. The quality parameter (0-100) used for compression varied between the slides. All WSIs were structured as tiled image pyramids with tile size 240$\times$240 or 256$\times$256 pixels\hl{ }and were saved in the SVS file format. Each included organ was represented by three slides, thus ensuring an organ-balanced dataset. The creation of a representative and diverse dataset was important for robust training of the deep learning models and for precise evaluation. Therefore, the WSIs were chosen to include varying image qualities, HE stain variations, and different artifacts. For example, some slides contained blurred regions or dust. The WSIs were also chosen to include various structures, textures and spatial distributions of tissue areas. In addition, particular care was taken to include WSIs with different amounts of glass and fat. All WSIs were quality checked by an experienced pathologist. 

The 21 WSIs constituted the raw dataset and were used for tissue segmentation and glass removal, and for the tissue compression task. For each task, a task-specific dataset was generated from these 21 WSIs. The tissue segmentation and glass removal was evaluated on whole image pyramids (Section \ref{sec:methodTissueSegAndGlassRem}), whereas the deep learning-based tissue compression was only evaluated on tissue patches (Section \ref{sec:methodTissueComp}). One deep learning model was evaluated on all patches from the whole image pyramids to give an estimate of the size reduction on image pyramid level, and comparison with JPEG and JPEG-XL compressed WSIs (Section \ref{sec:methods_glassRemovalAndTissueCompression}).

\subsection{Tissue segmentation and glass removal}\label{sec:methodTissueSegAndGlassRem}
\subsubsection{Tissue segmentation dataset}
Tissue segmentation was performed using a lightweight convolutional network. Feeding a neural network with an entire image pyramid, or a full image at high magnification, is not feasible, and a patch-wise approach was thus necessary. 

The tissue areas in all 21 slides were first delineated in QuPath~\cite{BankheadPeter2017QOss}. Since WSIs often contain complex structures, small fragments of tissue surrounded by glass were not annotated and small glass regions within large tissue areas were annotated as tissue to decrease annotation time. To reduce the burden of manually annotating huge WSIs, we used active learning. After a subset of the WSIs was annotated, an initial segmentation model was trained on this subset and used to generate annotations for the non-annotated data. The new annotations were then revised and corrected manually. Several iterations of this process were performed until the final ground truth segmentation masks were exported to the OME-TIFF format. 

From the 21 WSIs and corresponding ground truth annotation masks, patches with a size of $512\times512$ pixels were generated using FAST's PatchGenerator\cite{SmistadErik2015Fffh,SmistadErik2019HPNN}. Patches were extracted without overlap at the following magnifications: $\times$1.25, $\times$2.5, $\times$5, $\times$10, $\times$20 and $\times$40. Patches from magnification $\times40$ were exclusively extracted from WSIs scanned at $\times40$. As higher magnifications have larger image sizes than lower magnifications, the number of generated patches increased with approximately a fourfold when magnification doubled. Thus, more patches were extracted at magnifications $\times20$ and $\times40$ than at the lower magnifications. To keep the total amount of patches manageable, only every fourth of the possible $\times$20 and every eighth of the possible $\times$40 patches were included in the final dataset. For all patches, ground truth segmentations were retrieved from the annotations. Each HE-patch and its corresponding mask were then saved in the HDF5 format. The final dataset for tissue segmentation included more than 55 000 patches (Supplementary material Table 2).

\subsubsection{Tissue segmentation model}\label{sec:method_tissueSeg}
The tissue segmentation model was based on a U-Net architecture~\cite{ronneberger2015u} consisting of a contracting and an expanding path. Both were composed of five repeated blocks. Each contracting block contained one convolution block and a $2 \times 2$ max pooling layer. A convolution block was defined as two $3 \times 3$ convolutional layers, each followed by a batch normalization layer and a rectified linear unit (ReLU). The number of filters in the convolutional layers of the five convolution blocks were \{32, 32, 64, 128, 128\}. At the bottom of the U-Net, a convolution block with 128 filters connected the contracting and expanding paths. Each expanding block comprised an upsampling layer, a concatenation with the feature maps from the corresponding level in the contracting path, and two $3 \times 3$ convolutional layers, each followed by a ReLU. The first two expanding blocks also contained a batch normalization layer after each convolutional layer. This lightweight network had approximately 2 million parameters.

\subsubsection{Training}
Training was performed with TensorFlow v2.18.0~\cite{tensorflow2015-whitepaper} using the AdamW optimizer (Adam optimizer with decoupled weight decay)~\cite{loshchilov2019decoupledweightdecayregularization} with a learning rate of 1e$^{-4}$. Dice was used as the loss function, excluding the background. The maximum number of epochs was set to 300, and early stopping was implemented with 20 epochs as patience. 

The following data augmentation techniques were applied during training: horizontal and vertical flip, Gaussian blur with standard deviation of range $[0, 2]$, average pooling with kernel size $2 \times 2$, brightness variation with a factor in the range $[0.9, 1.1]$ and added value between -15 and 15, hue and saturation variation with a factor in the range $[0.9, 1.1]$~\cite{imgaug}. Furthermore, augmentation aiming to artificially reproduce glass-line artifacts was added to the glass-only patches (Fig. \ref{fig:overview_proposedMethod}h shows a real, unsegmented, glass-line artifact at the bottom), using the BlendAlphaRegularGrid imgaug method~\cite{imgaug}. A patch was split into a grid of 30 columns and 30 rows, where some rows and some columns were randomly selected following a binomial distribution. Rows were selected with probability $p = 0.004$, and columns were selected with probability $p = 0.004$. Their pixel values were then multiplied by a factor of 0.4, resulting in a gray appearance.

To sample equally from each magnification, a sampling approach was used during training and validation. Patches were randomly picked across the different magnifications, with each magnification having an equal probability of being sampled. Sampling was done without replacement and repeated every epoch. 

The tissue segmentation model was trained and assessed with five-fold cross-validation on slide level. All slides were randomly split into five folds, with each body part being represented with at most one WSI per fold. For each split, the segmentation model was trained on three folds, validated on the validation fold and evaluated on the WSIs in the test fold. The segmentation results for all five test folds were then combined. To evaluate the effect of magnification, the model was inferred at magnification $\times$1.25, $\times$2.5, $\times$5, $\times$10 and $\times$20. Throughout this experiment, patches were generated with 10\% overlap during inference to limit segmentation artifacts at the image patch edges.

The performance of the tissue segmentation model was evaluated quantitatively on all 21 WSIs, and compared quantitatively and qualitatively to the simple color thresholding algorithm in FAST~\cite{SmistadErik2015Fffh,SmistadErik2019HPNN}~\cite{pedersen2021fastpathology}. The color thresholding method segments at the image level with lowest resolution. As glass areas in WSIs generally appear white, pixels are classified as tissue or glass depending on their Euclidean distance to the color white. The distance is computed in RGB space as follows:
\begin{equation}
\begin{split}
    d &= \sqrt{(R-255)^2 + (G - 255)^2 + (B-255)^2}\\
    \label{eq:thresholding}
\end{split}
\end{equation}
where $R$, $G$, $B$ represent the \hl{intensities of the }red, green, and blue components of the pixel color \hl{with values between 0 and 255}, and the white color being represented as $(255, 255, 255)$ in RGB space. If the distance is above a given threshold (85 in our case), the pixel is classified as tissue, otherwise it is considered glass. The thresholding is followed by a morphological closing comprising dilation and erosion operations, both with a radius of nine.

\subsubsection{Inference}  
During inference, the image at a chosen magnification was separated into small patches that were individually processed by the network. The output of each patch was then stitched together to form the final segmentation at that magnification. As this may result in a large number of patches, especially for high magnifications, the processing runtime of a single patch needed to be kept low. A lightweight network performing fast patch inference was thus required. Tissue segmentation is not a very complex task and thus a lightweight network was considered to be sufficient.

\subsubsection{Glass removal}\label{glassremovaMethod}
\begin{figure}[h!]
    \centering
    \includegraphics[width=0.7\linewidth]{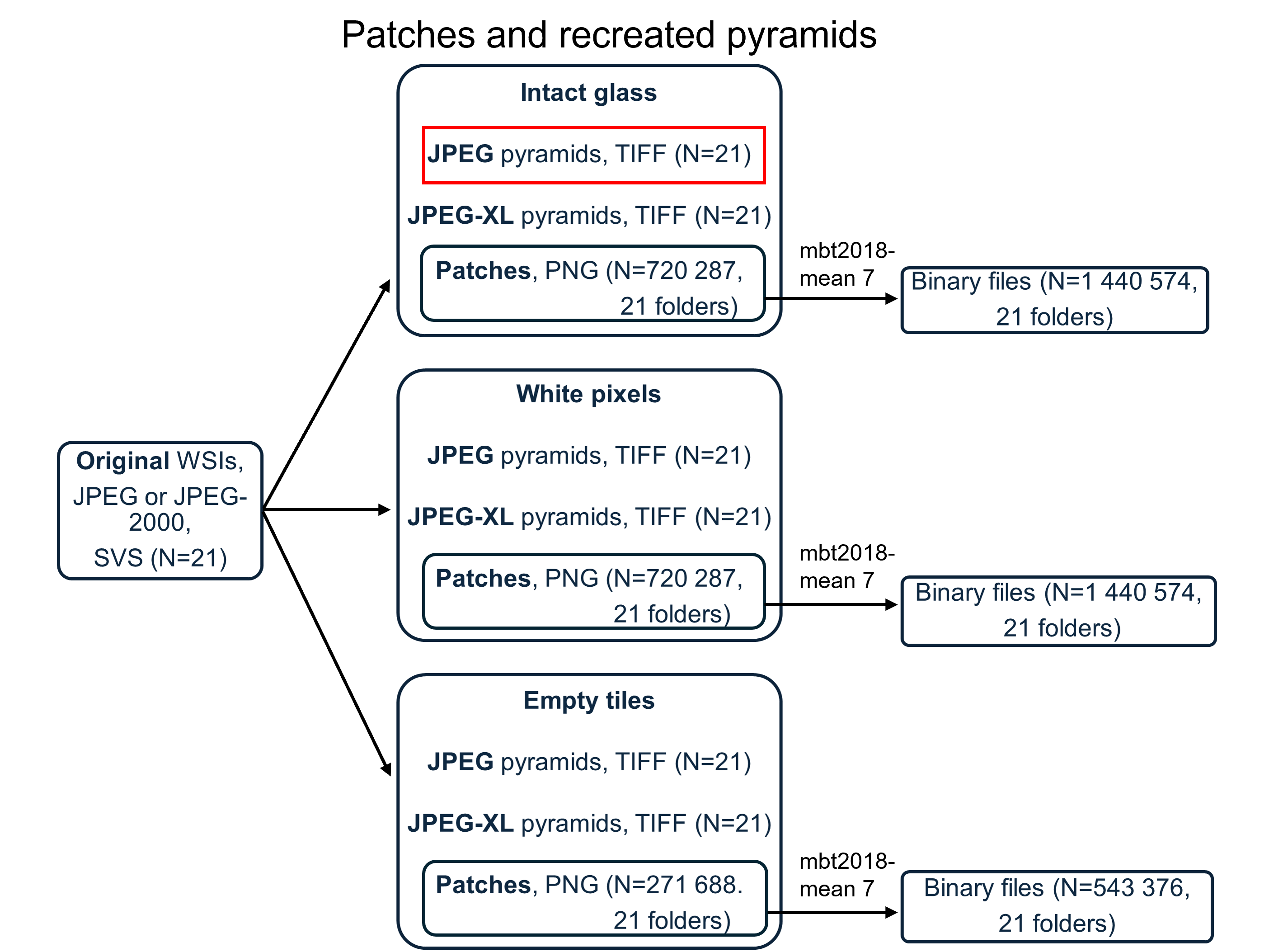}
    \caption{Creation of whole pyramids and \textit{patch pyramids}. JPEG-compressed (N=63) and JPEG-XL-compressed (N=63) image pyramids were created with the original SVS structure and tile size. Three image pyramids were created with each compression format: one with glass unchanged, one with glass pixels replaced by white pixels, and one with all-glass tiles removed. The recreated JPEG and JPEG-XL pyramids were compressed with quality level 90. The 21 original SVS image pyramids were also patched with size 256$\times$256 pixels into three \textit{patch pyramid} sets for each image: 1) An "original" patch set, where the glass was intact, 2) a "white" patch set, where glass was replaced by white pixels, 3) an "empty" patch set where the glass tiles were removed, and the glass pixels in tiles with glass and tissue were replaced by white pixels. The AI model \textit{mbt2018-mean 7} was inferred on the patches compressing them to bytestrings and saved as binary files. The compression rates were calculated by comparing the size of the image pyramids, and the combined size of all the binary files from each image to the size of the corresponding recreated JPEG-compressed pyramid with intact glass (outlined in red).}
    \label{fig:dataProcessing}
\end{figure}

Using the tissue segmentation model, pixels comprising glass were identified and removed from the image pyramids as illustrated in Fig.~\ref{fig:overview_proposedMethod}. As previously described, such glass areas do not provide any diagnostic information, hence are a waste of storage space. To construct new WSIs without glass, routines for creating image pyramids using the TIFF file format by libtiff were implemented in FAST~\cite{SmistadErik2015Fffh,SmistadErik2019HPNN}.
In this study, glass was removed from the WSIs in two different ways:\\
\\
\noindent\textbf{Single color method:} The first method consisted of replacing all pixels with pixels of a single color (e.g. white).
This may save space if the background is uneven or has artifacts. The benefit of this method is that it can be read by all WSI TIFF readers. However, images where all pixels have the same color are not compressed efficiently with JPEG.
For instance, a JPEG-compressed image of size $512\times512$ pixels containing only white pixels consumes over 4 kB (1 kB = 1024 bytes) of storage space.\\
\\
\noindent\textbf{Patch removal method:} The second method was to remove the all-glass tiles from the WSIs (Fig. \ref{fig:overview_proposedMethod}).
This can be achieved with the TIFF file format by storing the tiles as zero-byte tiles.
The benefit of this method is a reduced file size compared to the \textit{Single color method}, the disadvantage is that the WSI TIFF reader needs special routines to interpret these zero-byte tiles. This functionality was implemented in FAST as part of this study, and is also supported in OpenSlide~\cite{GoodeAdam2013OAvs} from version 4.0.0~\cite{openslideGenericTiFFformat}.
With this method, tiles that contain both tissue and glass would still need glass pixels to be converted to a single color. Tiles with only glass could be removed.\\
\\
From each WSI (original SVS file), six new image pyramids were created (Fig. \ref{fig:dataProcessing}). Three image pyramids were created with JPEG, and three with JPEG-XL compression. The pyramidal structure, and tile size from the original SVS files were kept, but all were compressed and saved with quality factor 90. The three image pyramids created for each image in each compression format were: one with\hl{ }unchanged\hl{ glass}, one with glass pixels replaced by white pixels, and one with all-glass tiles replaced by zero-byte tiles (Fig. \ref{fig:dataProcessing}). 

During the evaluation process, all image file sizes were calculated on the full image pyramids. The file size reduction was computed from the ratio between the size of the new image pyramid and the size of the corresponding recreated JPEG-compressed image pyramid with intact glass.

The preprocessing, compression, and image pyramid construction of the WSIs were implemented using FAST~\cite{SmistadErik2015Fffh,SmistadErik2019HPNN} version 4.1\hl{5}, libjxl 0.11.0 and Python 3.10. The pyramids were recreated on a single machine with an Intel Core i7 central processing unit (CPU) with 10 cores and 16 GB of RAM.

\subsection{Tissue compression}\label{sec:methodTissueComp}

The CompressAI framework~\cite{begaint2020compressai} was used to train and evaluate several different deep learning models for image compression of tissue areas in WSIs.
These models were then compared to non-machine learning methods for image compression (JPEG, JPEG-2000 and JPEG-XL).

\subsubsection{Data splits}
The 21 WSIs were randomly separated into a train (70\%), validation (15\%) or test (15\%) set (Supplementary material Table 1 and Table 3). The WSIs were patched with patch size $256\times256$ pixels using FAST's PatchGenerator~\cite{SmistadErik2015Fffh,SmistadErik2019HPNN}, and the generated patches were then compressed and saved with the lossless PNG, \textit{lossy} JPEG, JPEG-XL (libjxl v.0.11.0), and JPEG-2000 (openjpeg v.2.5.3) compression methods. The stored PNGs contained an alpha channel, whereas the rest did not.
For JPEG, the quality parameter was set to 90, and for JPEG-XL the distance quality parameter was set to be similar to JPEG quality of 90 using the \textit{JxlEncoderDistanceFromQuality} function in libjxl which gives a distance value of 1.0 and is currently the default value in libjxl. \hl{The compression effort for JPEG-XL was set to 7, which is the default value in libjxl.} For JPEG-2000, the peak signal-to-noise ratio (PSNR) quality value was set to 37 after being adjusted through experimentation to achieve a structural similarity index (SSIM) similar to that of JPEG-XL. The parameters for JPEG, JPEG-XL, and JPEG-2000 were kept throughout the paper.

Each WSI in the train and validation sets were patched from one of their existing image levels. The image level was chosen randomly from the WSI's existing image levels. Patches were only extracted from within the tissue areas, which were identified with FAST's color threshold tissue segmentation method, and only patches with more than 50\% tissue were included~\cite{SmistadErik2019HPNN}. If a very high level was selected, the size of the level could be very small, which could result in zero patches being extracted. To prevent this, a cut off on segmented tissue area was set. If fewer than $2\times512\times512$ pixels were identified as tissue with the tissue segmentation method at the chosen level, level zero was automatically selected. The resulting train and validation sets consisted of 52 294 and 1 707 patches, respectively (Supplementary material Table 3). 

The WSIs in the test set were also patched from within the tissue area. However, in the test set, tissue regions were identified by the proposed tissue segmentation method trained on the one-fold data split used in the tissue compression task (Supplementary material Table 3). From these WSIs, all possible patches were extracted at magnification $\times$1.25, $\times$2.5, $\times$5, $\times$10, $\times$20 and $\times$40, creating six separate test sets with 138, 565, 2 296, 9 168, 36 947, and 6 045 patches, respectively (Supplementary material Table 4). Then 138 patches were randomly selected from each magnification ($\times$1.25, $\times$2.5, $\times$5, $\times$10, $\times$20, $\times$40), to generate an additional balanced test set of 828 patches.

\subsubsection{Network architectures}\label{Tissue_comp_metrics_protocols}
The pretrained compression models bmshj2018-factorized~\cite{BalléJohannes2018Vicw}, bmshj2018-hyperprior~\cite{BalléJohannes2018Vicw} and mtb2018-mean~\cite{MinnenDavid2018JAaH}, implemented and available in CompressAI~\cite{begaint2020compressai}, were downloaded and evaluated on the balanced test set without modification or further training by us. The models were available at different quality levels (1-8), corresponding to different lambda values in the rate-distortion loss, as well as having different channel numbers in the entropy bottleneck and in the convolutional layers for some of the models' quality levels~\cite{compressAIModelZoo}. The models were evaluated at three quality levels, corresponding to three rate-distortion trade offs in the loss function (quality level/lambda values of 5/0.0250, 6/0.0483, and 7/0.0932)~\cite{compressAIModelZoo}. 
In addition, the compression models bmshj2018-factorized~\cite{BalléJohannes2018Vicw}, bmshj2018-hyperprior~\cite{BalléJohannes2018Vicw}, mtb2018-mean~\cite{MinnenDavid2018JAaH}, and Elic2022Chandelier (based on the context model and attention model by He \textit{et al.}~\cite{HeDailan2022EELI} and Cheng \textit{et al.}~\cite{ChengZhengxue2020LICw}, respectively, and parameters and simplifications by Chandelier~\cite{chandelierV2023}), implemented and available in CompressAI~\cite{begaint2020compressai}, were trained on patches from histopathology images with and without augmentation.
This was done to evaluate whether better compression could be achieved by training on domain-specific pathology images.

\subsubsection{Training}
The models trained on pathology data were trained from scratch over 130 epochs, with the Adam optimizer and an initial learning rate of 1e$^{-4}$. The loss function was a rate distortion loss~\cite{begaint2020compressai}, either optimized for the mean squared error (MSE) or SSIM as the distortion loss. A lambda value determined the weight of the distortion loss. The models were trained with and without augmentation. The following data augmentation techniques were applied during training, each with a 50\% probability: Gaussian blur, ColorJitter (brightness, hue, saturation, contrast), horizontal flip, and vertical flip.

The patches were compressed to bytestrings. The compression was evaluated by the average number of bits per pixel (bpp) of the bytestrings, and as the size (kB) of the bytestrings. 
The average amount of saved space per patch was calculated as the ratio between the size of the compressed patches (kB) and the size of the corresponding JPEG patch (Supplementary material, Eq. 1). The image quality was evaluated with the image quality metrics SSIM and PSNR between the original PNG patch and the decompressed patch after inference. Decompression time was estimated as the average decompression time per patch, ignoring the first 10 patches. Each metric was reported as the average score for a patch in the test set.

All models were evaluated on the balanced test set. To evaluate the effect of magnification, the pretrained models at quality level seven and corresponding trained models (without augmentation) were also evaluated on the six magnification test sets. 

Preprocessing of WSIs was done with FAST~\cite{SmistadErik2015Fffh,SmistadErik2019HPNN}, experiments were performed with Python v3.10, and model training with PyTorch v2.5.1.~\cite{paszke2019pytorch} and CompressAI v1.2.6~\cite{begaint2020compressai}. 

\subsubsection{Experiments}
\noindent\textbf{Value of training on histopathology data}:
The pretrained compression models, bmshj2018-factorized~\cite{BalléJohannes2018Vicw}, bmshj2018-hyperprior~\cite{BalléJohannes2018Vicw} and mbt2018-mean~\cite{MinnenDavid2018JAaH}, optimized for MSE in the rate-distortion loss with lambda values of 0.0250, 0.0483 and 0.0932 were evaluated on the test set. 
To evaluate the effect of domain-specific training, the same networks were also trained on histopathology data.
Table \ref{tab:experiments} shows all the compression models trained and evaluated.\\

\begin{table}[h!]
    \centering
    \begin{tabular}{l l c c c r}
    \toprule
         Pre & Arch & $\mathcal{D}$ Loss & $\lambda$ & Aug & No Aug\\
         \midrule
         Y & factorized & MSE & 0.0250  & &\\
               &                      &     & 0.0483  &  &\\
               &                      &     & 0.0932  & &\\         
               & hyperprior & MSE & 0.0250  & \\
               &                      &     & 0.0483  & &\\
               &                      &     & 0.0932  & &\\        
               & mbt2018         & MSE & 0.0250  & \\
               &                      &     & 0.0483  & &\\
               &                      &     & 0.0932  & &\\
        \midrule
         N & factorized & MSE & 0.0250 & \cmark & \cmark\\
            &                  &     & 0.0483 & \cmark & \cmark\\
            &                  &     & 0.0932 & \cmark & \cmark\\
            &                  & SSIM & 31.73 & \cmark & \cmark\\
            &                  &      & 60.50 & \cmark & \cmark\\
        & hyperprior & MSE & 0.0250 &  \cmark & \cmark\\
          &                   &     & 0.0483 & \cmark & \cmark\\
          &                   &     & 0.0932 & \cmark & \cmark\\
          &                   & SSIM & 31.73 & \cmark & \cmark\\
          &                   &      & 60.50 & \cmark & \cmark\\
        & mbt2018         & MSE & 0.0250 & \cmark & \cmark\\
          &                   &     & 0.0483 & \cmark & \cmark\\
          &                   &     & 0.0932 & \cmark & \cmark\\
          &                   & SSIM & 31.73 & \cmark & \cmark\\
          &                   &      & 60.50 & \cmark & \cmark\\
        & Elic2022 & MSE  & 0.03 & \cmark &\\
          &                 &      & 0.045 &  \cmark &\\
          &                 & SSIM & 60.50   &  \cmark &\\
        \bottomrule
        
    \end{tabular}
    \caption{Pretrained models available in CompressAI~\cite{begaint2020compressai}, and models trained on pathology data which were evaluated in this study. Abbreviations: $\mathcal{D}$ = distortion, Aug = augmentation (trained with), No Aug = no augmentation (trained without), MSE = mean squared error, SSIM = structural similarity index, Pre. = pretrained (not trained further, used as is), Y = yes, N = no (trained from scratch), Arch = architecture, factorized = bmshj2018-factorized, hyperprior = bmshj2018-hyperprior, mbt2018 = mbt2018-mean, Elic2022 = Elic2022Chandelier}
    \label{tab:experiments}
\end{table}

\noindent\textbf{The effect of the distortion loss}:
Two distortion components in the rate-distortion loss were evaluated (Table \ref{tab:experiments}). The rate distortion loss was optimized with MSE or SSIM as the distortion component. The lambda values for the MSE optimized models were 0.0250, 0.03, 0.045, 0.0483 and 0.0932, whereas the lambda values for the SSIM optimized models were 31.73 and 60.50. \\
\\
\noindent\textbf{The effect of augmentation}:
The models bmshj2018-factorized, bmshj2018-hyperprior, mtb2018-mean were trained with and without augmentation (Table \ref{tab:experiments}). When augmentation was applied, each of the four augmentations had a 50\% probability of being added.\\
\\
\noindent\textbf{Compression rate on different magnification levels}:
Since WSIs typically include downsampled versions of the same image, a tissue compression method needs to compress image patches at different magnifications. We observed a difference in compression rate between magnifications on our validation set. During testing, the pretrained and corresponding models without augmentation were thus evaluated on the six test sets at magnifications $\times$1.25, $\times$2.5, $\times$5, $\times$10, $\times$20, and $\times$40 to see if there was a difference in image quality.\\
\\
\noindent\textbf{Benchmark against non-learning compression methods}:
The deep learning models were compared with JPEG-XL and JPEG-2000 using the SSIM, PSNR, bpp and file size reduction metrics. \\
\\
\noindent\textbf{Worst case - artifacts}:
The patches with the lowest SSIM for the top performing deep learning models, JPEG-2000 and JPEG-XL were evaluated qualitatively to look for prominent artifacts\hl{ }and assess the models' suitability for clinical use. Even with high average quality metrics, models generating outliers with artifacts may need to be discarded to prevent potential errors in the clinic.

For these patches, the decompressed deep learning patch, JPEG-2000 patch and JPEG-XL patch were converted to gray scale through averaging the RGB channels of the images, and then subtracted from the gray scale version of its corresponding PNG patch to create a difference map which visualizes errors introduced by the compression.

\subsection{Glass removal and tissue compression}\label{sec:methods_glassRemovalAndTissueCompression}
To enable comparison of the glass removal on JPEG and JPEG-XL-compressed pyramids with glass removal and AI compression, all 21 original SVS image pyramids were also patched with patch size 256$\times$256 pixels without overlap on all available image levels. Patches at the edge were padded with white. The generated patches were saved as PNG patches in per-image folders (\textit{patch pyramids)}. Patching was performed on the original SVS files to avoid double compression prior to AI compression. Three patch folders were created for each of the 21 images, one with patches from the original image, one where glass pixels were replaced with white pixels, and one where glass patches were removed. Glass was identified by inferring the segmentation model presented in Section \ref{sec:method_tissueSeg}, inferred on magnification $\times$2.5 with 10\% overlap and rescaled to each image level. The PNG-patches were then compressed with the pretrained mbt2018-mean model at quality level 7 optimized for MSE, and the bytestrings were saved as binary files on disk (Fig. \ref{fig:dataProcessing}). The combined size of all binary files for an image pyramid was calculated, and compared against the size of the corresponding recreated JPEG-compressed image pyramid with intact glass (red box in Fig. \ref{fig:dataProcessing}). All file sizes were calculated as the true file size, not allocated memory.

\section{Results}
\subsection{Tissue segmentation and glass removal}
\subsubsection{Tissue segmentation evaluation}
\begin{figure*}[ht]
    \centering
    \includegraphics[width=0.9\linewidth]{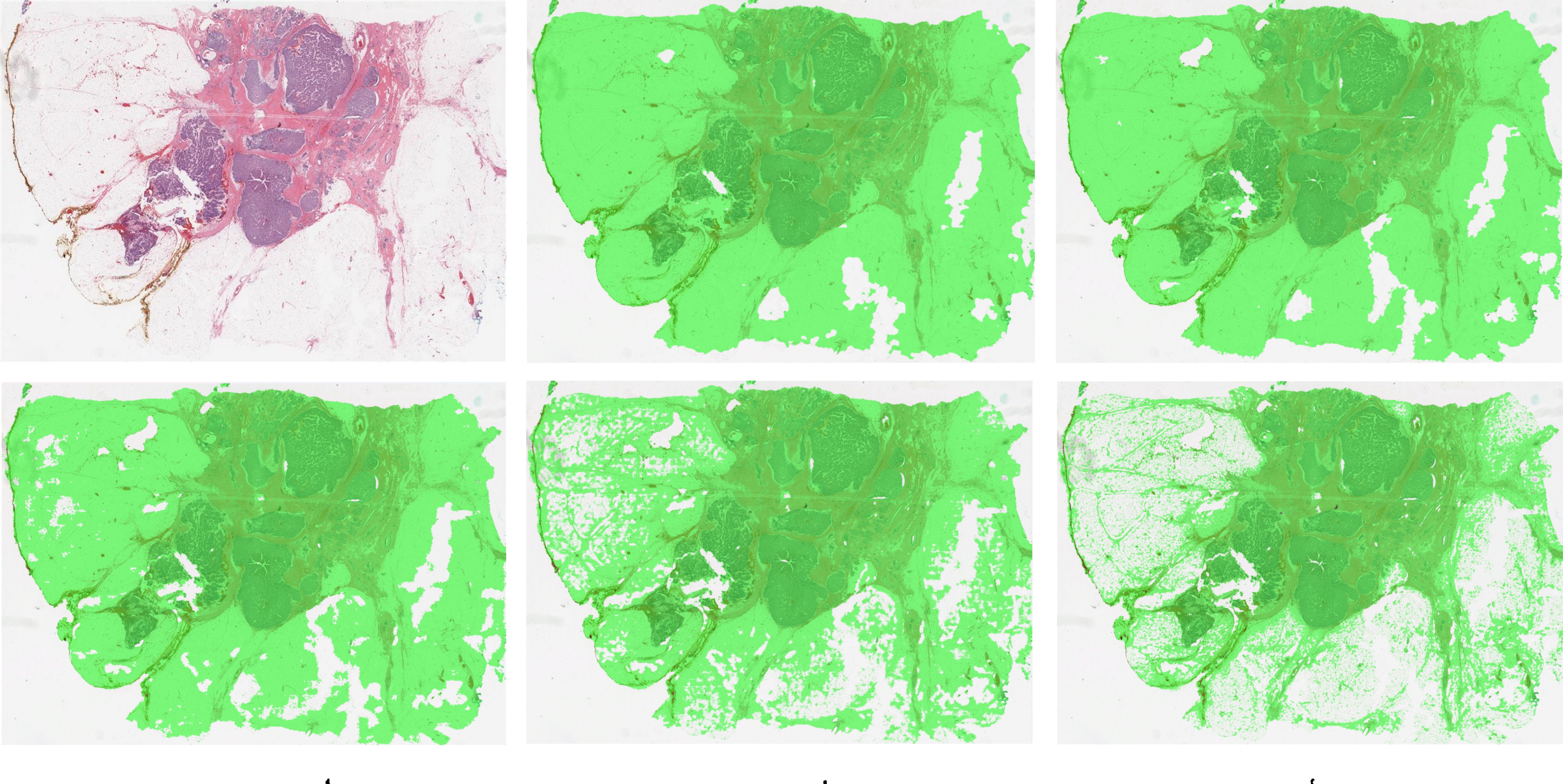}
    \caption{Tissue segmentation with the proposed model inferred on an image of a hematoxylin and eosin (HE)-stained section (top left) at magnification $\times$1.25 (top middle), $\times$2.5 (top right), $\times$5 (bottom left), $\times$10 (bottom middle), $\times$20 (bottom right). The adipose tissue areas seen as unstained, white tissue in the HE-stained section (top left) are segmented at magnification x1.25 (top middle). The inclusion of adipose tissue in the segmentations decreased with increased magnification. \hl{The false negative rate, i.e. the tissue loss, at $\times$1.25, $\times$2.5, $\times$5, $\times$10 and $\times$20 were 0.026, 0.012, 0.024, 0.165 and 0.370, respectively.} The WSI in this illustration is the Her2Neg\_Case\_22 from the HER2 tumor ROIs dataset~\cite{Farahmand22} available at The Cancer Imaging Archive (TCIA)\cite{clark2013cancer}, with a \href{https://creativecommons.org/licenses/by/4.0/}{CC BY 4.0} license, and has been modified for this illustration.}
    \label{fig:her2neg_case_22}
\end{figure*}

\begin{figure*}[ht]
    \centering
    \includegraphics[width=0.9\linewidth]{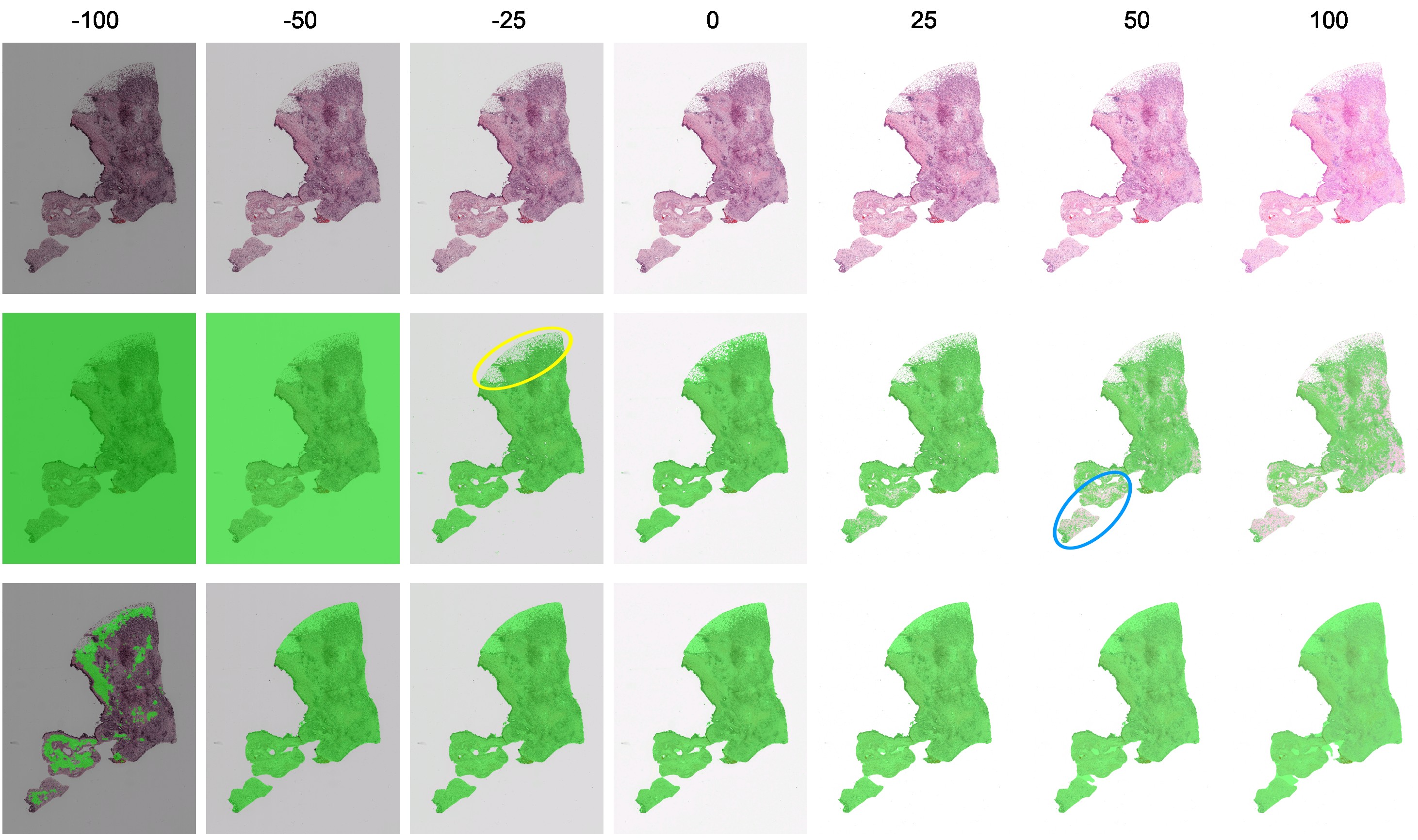}
    \caption{Tissue segmentation with color thresholding and the proposed model at multiple brightness levels. The top row represents a WSI with varying brightness factors (-100 to 100), brightness factor zero represents the original WSI. The WSI was converted to the HSV color space, and a value between -100 and 100 was added to the V channel. Row two shows tissue segmentation with the color thresholding algorithm inferred at the highest available image level on the seven different brightness variations. Row three shows the proposed neural network segmentation model inferred at magnification $\times$2.5 with 10\% patch overlap. In row two, the color thresholding algorithm struggles with segmenting fat (yellow circle) and light colored areas (blue circle). The figure illustrates that tissue segmentation with the neural network is more robust to brightness variations than a color thresholding algorithm. The WSI in this illustration is the C3L-02897-21 from the CPTAC-PDA dataset~\cite{CPTAC-PDA18} available at The Cancer Imaging Archive (TCIA)\cite{clark2013cancer}, with a \href{https://creativecommons.org/licenses/by/3.0/}{CC BY 3.0} license, and has been modified for this illustration.}
    \label{fig:brightness}
\end{figure*}

Dice scores of 0.971-0.983 were achieved by the tissue segmentation model on the 21 slides at the five different magnifications (Table \ref{tab:resultsTissueSeg}). Differences in segmentation performance were observed between magnifications during qualitative assessment. At high magnifications, the model excluded the inside of fat cells in some WSIs (Fig. \ref{fig:her2neg_case_22}). Adipose tissue may be seen as the white, unstained tissue areas in Fig. \ref{fig:her2neg_case_22}, illustrating the challenge of distinguishing it from glass. In some WSIs, segmenting at high magnification resulted in the incorrect segmentation of artifacts as tissue. Using the manual annotations as ground truths, the color thresholding method achieved a Dice score of 0.954. When qualitatively comparing the tissue segmentation model with standard color thresholding on three random slides, at six brightness variations, the deep learning segmentation outperformed the color thresholding at the darkest and brightest cases (Fig. \ref{fig:brightness}). However, the proposed deep learning segmentation model also struggled at the darkest brightness variation (Fig. \ref{fig:brightness}), whereas the color thresholding model achieved very precise segmentations at some brightness variations.

\newcolumntype{Y}{>{\centering\arraybackslash}X}
\begin{table}[h!]
    \centering
    \begin{tabular}{l l c c}
        \toprule
         Method & Magnification & Dice & \hl{Runtime (s)} \\
         \midrule
         \multirow{5}*{Proposed model}& $\times$1.25 & 0.971 $\pm$ 0.044 & \hl{0.2 $\pm$ 0.2} \\
         & $\times$2.5 & 0.980 $\pm$ 0.034 & \hl{0.9 $\pm$ 1.0}\\
         & $\times$5 & 0.983 $\pm$ 0.029 & \hl{4.7 $\pm$ 3.8}\\
         & $\times$10 & 0.980 $\pm$ 0.027 & \hl{11.9 $\pm$ 9.7}\\
         & $\times$20 & 0.975 $\pm$ 0.044 & \hl{53.3 $\pm$ 41.2}\\
         \midrule
         Color thresholding & Lowest available & 0.954 $\pm$ 0.077 & \hl{0.2 $\pm$ 0.2}\\
         \bottomrule
    \end{tabular}
    \caption{Dice score \hl{and runtime (in seconds)} of the trained tissue segmentation model inferred on \hl{WSIs at} magnifications $\times1.25$, $\times2.5$, $\times5$, $\times10$ and$\times20$ with 10\% patch overlap. The last row corresponds to the color thresholding method. Average values with standard deviation indicated by $\pm$ over all 21 WSIs in the dataset.}
    \label{tab:resultsTissueSeg}
\end{table}

\subsubsection{File size reduction}
\begin{figure}[h!]
    \centering
    \includegraphics[width=0.5\linewidth]{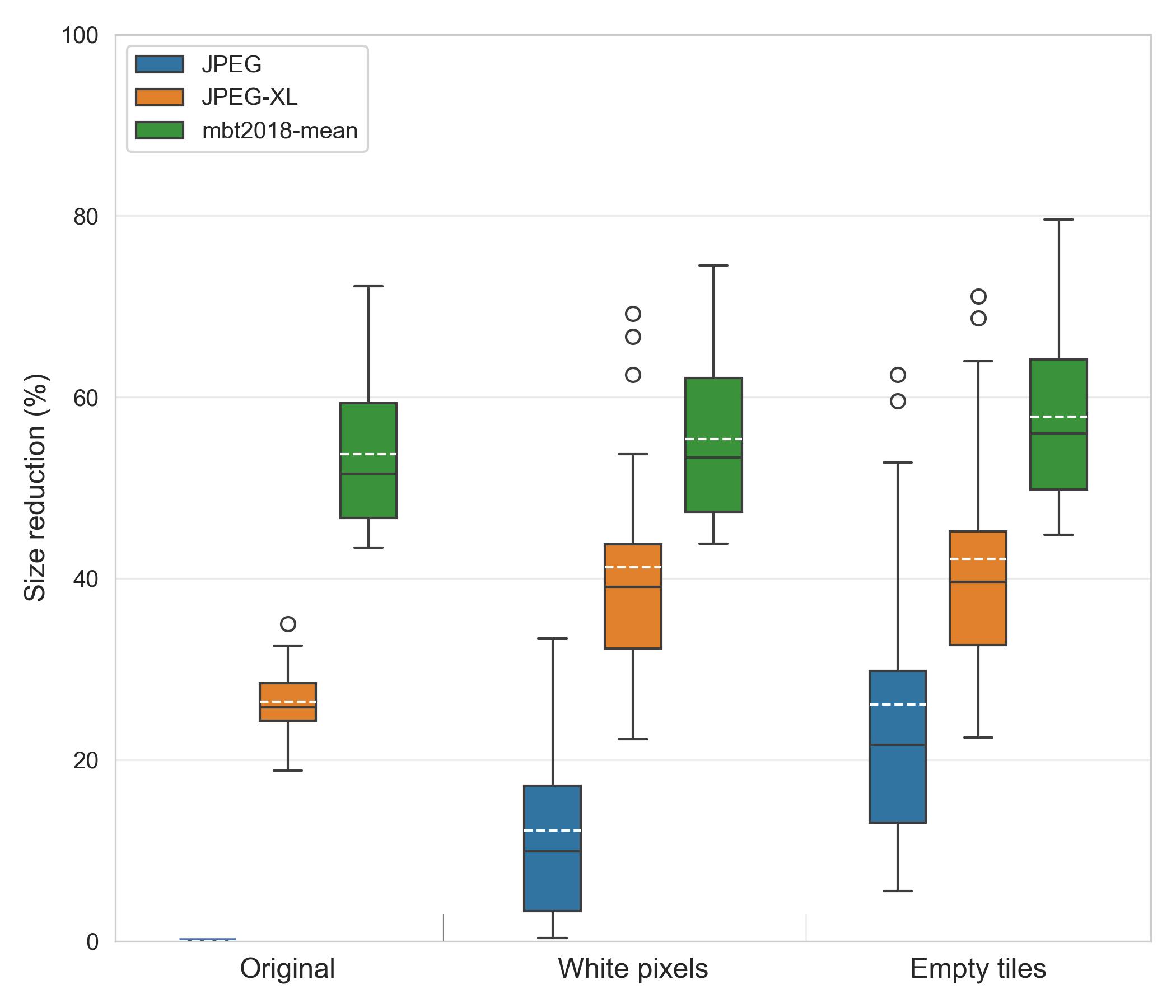}
    \caption{Boxplot representing the size reduction (\%) of new image pyramids and \textit{patch pyramids} compared to the original JPEG-compressed image pyramid. The tissue segmentation model was inferred at magnification $\times$2.5 with 10\% overlap. The recreated JPEG-compressed (white, empty) and JPEG-XL-compressed (original, white, empty) image pyramids, and the pretrained mbt2018-mean model at quality level 7 (original, white, empty) compressed \textit{patch pyramids} all saved space compared to the JPEG-compressed images with intact glass (original). The original columns represent the size reduction for JPEG, JPEG-XL and the pretrained mbt2018-mean model at quality level 7 with intact glass compared to the JPEG-compressed image pyramid. The white stapled lines represent the mean, and the black full line the median.}
    \label{fig:glassRemovalBoxplot}
\end{figure}

Replacing glass pixels with white pixels, and replacing glass tiles with empty tiles, saved space for both JPEG and JPEG-XL compression compared to their corresponding image pyramid with intact glass. JPEG-XL outperformed JPEG for the images with intact glass, the images where glass was replaced by white pixels, and for the images where glass tiles were replaced by empty tiles (Fig. \ref{fig:glassRemovalBoxplot}). 

After segmenting tissue at magnification $\times$2.5 with 10\% overlap, JPEG-compressed image pyramids achieved a size reduction between 0.3 and 33\%, and 6 and 62\%, when replacing glass with white pixels and empty tiles, respectively. JPEG-XL saved between 19 and 35\% on the images with intact glass. Replacement of glass with white pixels and JPEG-XL compression resulted in a size reduction between 22 and 69\%, whereas replacement of glass with empty tiles and JPEG-XL compression resulted in a size reduction between 22 and 71\% (Fig. \ref{fig:glassRemovalBoxplot}).

The total size of all 21 WSIs in the dataset was 6.45 GB when JPEG compression and no glass removal was performed. Replacing glass tiles with empty tiles and keeping JPEG-compression led to a total size for all WSIs of 5.07 GB, corresponding to a total size reduction of 22\%. By replacing glass tiles with empty tiles and using JPEG-XL compression, the total size of all WSIs dropped to 3.95 GB which corresponds to a total size reduction of 39\%. All file size reduction results are compared to the JPEG-pyramid with intact glass. 

\begin{figure}[h!]
    \centering
    \includegraphics[width=0.4\linewidth]{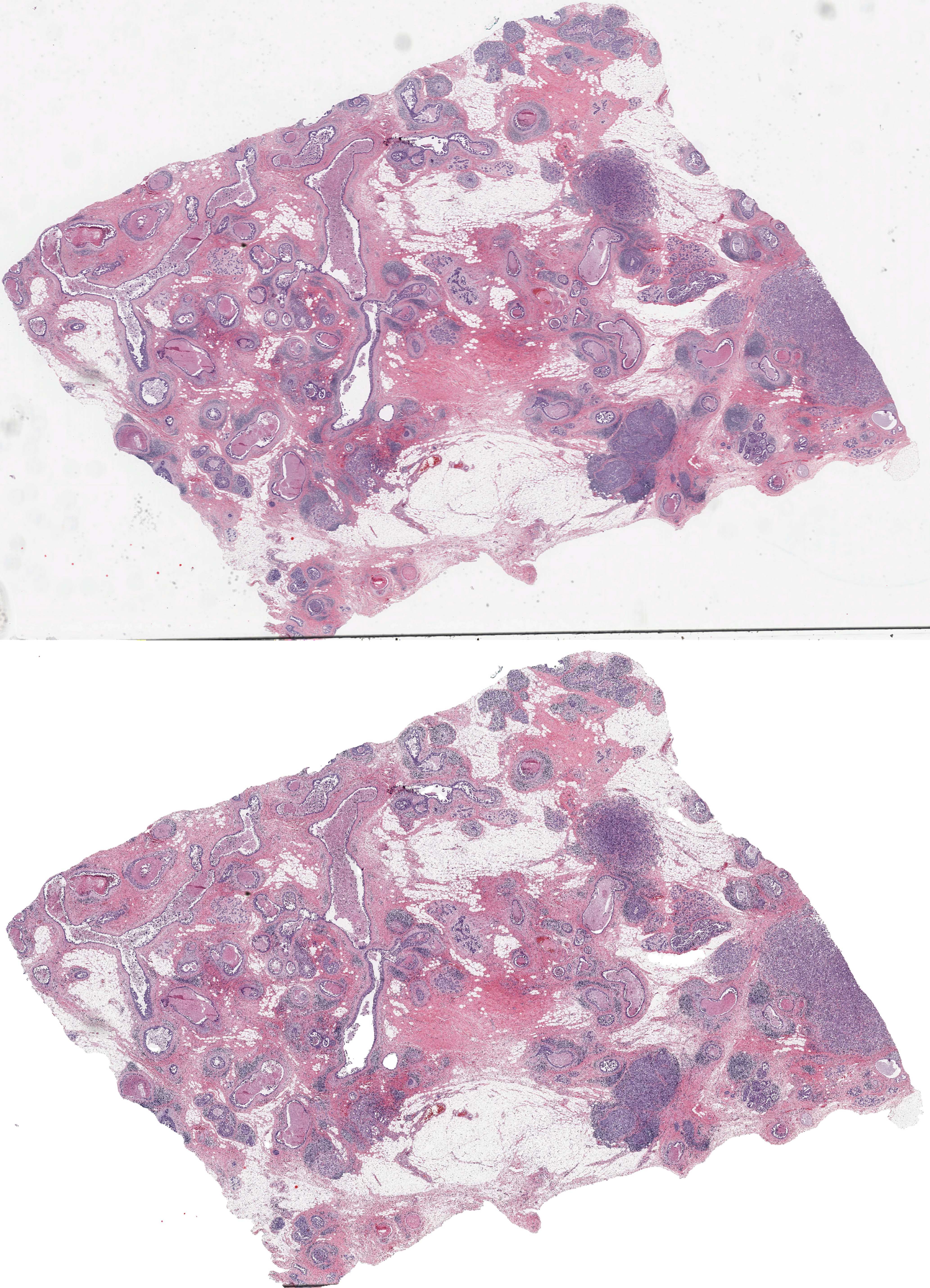}
    \caption{Original whole slide image (WSI) of a hematoxylin and eosin stained section from a breast tissue (top) and the same WSI where the glass tiles have been replaced by empty tiles (bottom) using the model inferred at $\times$2.5 and 10\% overlap. The top figure shows an example of the glass-line artifacts that may be present in WSIs (black horizontal line at bottom). The proposed tissue segmentation model was able to segment this correctly as glass and thus remove it from the image as shown in the bottom figure. The color thresholding method segmented the line incorrectly as tissue. The WSI in this illustration is the Her2Pos\_Case\_37 from the HER2 tumor ROIs dataset~\cite{Farahmand22} available at The Cancer Imaging Archive (TCIA)\cite{clark2013cancer}, with a \href{https://creativecommons.org/licenses/by/3.0/}{CC BY 3.0} license, and has been modified for this illustration.}
    \label{fig:glassRemoval}
\end{figure}

Fig. \ref{fig:glassRemoval} shows an original WSI and the corresponding JPEG-compressed image pyramid with the glass tiles replaced by empty tiles. Compared to the JPEG-compressed image pyramid containing the original tiles, the file size was reduced from 501 MB to 465 MB which represents a 7\% file size reduction, one of the lowest reductions in our dataset (1 kB = 1024 bytes). The glass-line artifact at the bottom of the image was not identified as tissue and removed.
The file size reduction by glass removal is highly dependent on the amount of glass in the image.

\clearpage

\subsubsection{Runtime}
\hl{The} runtime experiments were performed using a \hl{NVIDIA GeForce RTX 4060 (8GB)} and the TensorRT v8.6 inference engine \hl{on a Windows system}.
The average tissue segmentation runtime was \hl{15$\pm$3} ms for 512$\times$512 sized patches.
The runtime of the tissue segmentation on a WSI depends on the number of patches to process. The number of patches depends on several factors such as magnification level and patch overlap. In addition to inference runtime, processing a whole WSI also includes the time to generate patches and stitch the segmented patches to create a full segmentation mask. In this study, this was done using FAST's PatchGenerator and PatchStitcher.
\hl{The average tissue segmentation runtime per WSI, using 10\% patch overlap, is given in Table} \ref{tab:resultsTissueSeg}.

\hl{The total recompression runtime for the new JPEG-compressed image pyramids were on average 26, 39, and 39 seconds when including glass, replacing glass with white pixels, and replacing glass with blank tiles, respectively (Supplementary material, Table 5). The total recompression runtime for the new JPEG-XL-compressed image pyramids were on average 835, 926, and 344 seconds when including glass, replacing glass with white pixels, and replacing glass with blank tiles, respectively (Supplementary material, Table 5). A full runtime breakdown can be found in Table 5 in the Supplementary material for recompression of WSIs.}

\subsection{Tissue compression}
Table~\ref{tab:results_tissueCompt} shows the compression results on the tissue patches in the balanced test set, both in terms of image quality and file size reduction of the compression models and algorithms tested in this study.
On the balanced test set, the deep learning models could on average reduce the file size of a tissue patch by up to 70\% compared to JPEG compression. However, these models resulted in a low image quality with an average SSIM of about 0.90.
The deep learning models with the highest average SSIMs (above 0.95), comparable to that of JPEG-XL and JPEG-2000, saved approximately 34-40\% on average per patch compared to JPEG (bmshj2018-hyperprior and mbt2018-mean). The bmshj2018-hyperprior and mbt2018-mean models, with lambda 0.0932 for the MSE optimized models and 60.5 for the SSIM optimized models, achieved the highest SSIM scores. \\

\begin{table*}
    \begin{adjustbox}{width=\textwidth}
    \begin{tabular}{rlrllllm{1.7cm}rm{2.5cm}rm{1cm}}
    \toprule
    Pre. & Arch. & Q & $\lambda$ & $\mathcal{D}$ loss & aug & SSIM (min, max) & PSNR (min, max) & total [kB] & \hl{Size reduction} (min, max) [\%] & bpp & dec time [ms] \\
    \midrule
     & JPEG &  &   &  &   & 0.95 (0.87, 1.0) & 35 (28, 54) & 18678 & 0 (0, 0) & 2.82 &  2 \\
     & JPEG-2000 &  &   &  &   & 0.96 (0.9, 0.99) & 37 (28, 47) & 18812 & 14 (-74, 92) & 2.84 & - \\
     & JPEG-XL &  &   &  &   & 0.96 (0.9, 1.0) & 37 (28, 54) & 15293 & 17 (-3, 60) & 2.31 &  3\\
    \midrule
    1 & factorized & 5 &   & MSE &   & \textcolor{blue}{\textbf{0.89 (0.68, 1.0)}} & 32 (21, 49) & 5309 & \textcolor{blue}{\textbf{71 (14, 78)}} & 0.80 & 10 \\
     & factorized & 6 &   & MSE &   & 0.92 (0.8, 1.0) & 33 (24, 50) & 8187 & 56 (-18, 67) & 1.24 & 18 \\
     & factorized & 7 &   & MSE &   & 0.95 (0.85, 1.0) & 35 (26, 51) & 11624 & 38 (-62, 52) & 1.75 & 20 \\
     & hyperprior & 5 &   & MSE &   & \textcolor{blue}{\textbf{0.91 (0.78, 0.99)}} & 33 (24, 48) & 5882 & \textcolor{blue}{\textbf{70 (62, 86)}} & 0.89 & 19 \\
     & hyperprior & 6 &   & MSE &   & 0.94 (0.86, 1.0) & 34 (27, 50) & 8649 & 56 (39, 83) & 1.31 & 23 \\
     & hyperprior & 7 &   & MSE &   & \textbf{0.96 (0.9, 0.99)} & 36 (29, 49) & 11452 & \textbf{41 (26, 80)} & 1.73 & 24 \\
     & mbt2018 & 5 &   & MSE &   & 0.92 (0.8, 1.0) & 33 (27, 48) & 6267 & 69 (51, 91) & 0.95 & 26 \\
     & mbt2018 & 6 &   & MSE &   & 0.94 (0.86, 1.0) & 35 (28, 52) & 8580 & 58 (39, 90) & 1.30 & 26 \\
     & mbt2018 & 7 &   & MSE &   & \textbf{0.96 (0.9, 1.0)} & 36 (29, 53) & 11510 & \textbf{42 (24, 86)} & 1.74 & 26 \\
    \midrule
    0 & factorized & 5 & 0.025 & MSE & no & 0.88 (0.66, 0.99) & 31 (20, 47) & 5377 & 68 (-54, 77) & 0.81 & 10 \\
     & factorized & 5 & 0.025 & MSE & yes & 0.87 (0.63, 0.99) & 31 (21, 49) & 5085 & 70 (-37, 79) & 0.77 & 10 \\
     & factorized & 5 & 31.73 & SSIM & no & 0.91 (0.74, 1.0) & 31 (21, 44) & 6934 & 57 (-122, 74) & 1.05 & 10 \\
     & factorized & 5 & 31.73 & SSIM & yes & 0.91 (0.74, 1.0) & 31 (21, 44) & 7052 & 57 (-87, 72) & 1.06 & 10 \\
     & factorized & 6 & 0.0483 & MSE & no & 0.9 (0.65, 0.99) & 32 (19, 50) & 7914 & 53 (-116, 68) & 1.19 & 17 \\
     & factorized & 6 & 0.0483 & MSE & yes & 0.9 (0.72, 1) & 32 (22, 49) & 7552 & 56 (-88, 68) & 1.14 & 18 \\
     & factorized & 6 & 60.5 & SSIM & no & 0.94 (0.84, 1.0) & 32 (19, 47) & 10710 & 33 (-238, 61) & 1.62 & 15 \\
     & factorized & 6 & 60.5 & SSIM & yes & 0.94 (0.83, 1.0) & 32 (23, 45) & 10512 & 36 (-173, 60) & 1.59 & 16 \\
     & factorized & 7 & 0.0932 & MSE & no & 0.93 (0.8, 1.0) & 34 (22, 50) & 11561 & 32 (-209, 50) & 1.75 & 20 \\
     & factorized & 7 & 0.0932 & MSE & yes & 0.93 (0.79, 1.0) & 33 (23, 51) & 10886 & 37 (-152, 55) & 1.64 & 19 \\
     & hyperprior & 5 & 0.025 & MSE & no & 0.9 (0.41, 0.99) & 32 (11, 48) & 5827 & 69 (61, 85) & 0.88 & 18 \\
     & hyperprior & 5 & 0.025 & MSE & yes & 0.9 (0.76, 1.0) & 32 (23, 49) & 5615 & 71 (64, 85) & 0.85 & 20 \\
     & hyperprior & 5 & 31.73 & SSIM & no & 0.93 (0.81, 1.0) & 31 (21, 46) & 7650 & 57 (26, 75) & 1.15 & 20 \\
     & hyperprior & 5 & 31.73 & SSIM & yes & 0.93 (0.81, 1.0) & 31 (23, 45) & 7464 & 58 (22, 76) & 1.13 & 20 \\
     & hyperprior & 6 & 0.0483 & MSE & no & 0.93 (0.79, 0.99) & 34 (21, 49) & 8791 & 54 (42, 83) & 1.33 & 23 \\
     & hyperprior & 6 & 0.0483 & MSE & yes & 0.93 (0.84, 1.0) & 33 (25, 48) & 8411 & 56 (45, 84) & 1.27 & 23 \\
     & hyperprior & 6 & 60.5 & SSIM & no & \textbf{0.96 (0.81, 1.0)} & 33 (18, 46) & 11354 & \textbf{35 (-17, 56)} & 1.71 & 24 \\
     & hyperprior & 6 & 60.5 & SSIM & yes & \textbf{0.96 (0.88, 1.0)} & 33 (25, 45) & 11175 & \textbf{37 (-18, 59)} & 1.69 & 24 \\
     & hyperprior & 7 & 0.0932 & MSE & no & 0.95 (0.87, 1.0) & 35 (23, 51) & 11830 & 37 (20, 81) & 1.79 & 24 \\
     & hyperprior & 7 & 0.0932 & MSE & yes & \textbf{0.95 (0.87, 1.0)} & 35 (26, 50) & 11513 & \textbf{39 (15, 81)} & 1.74 & 24 \\
     & mbt2018 & 5 & 0.025 & MSE & no & 0.9 (0.77, 0.99) & 32 (22, 46) & 6110 & 69 (-255, 95) & 0.92 & 26 \\
     & mbt2018 & 5 & 0.025 & MSE & yes & 0.9 (0.78, 1.0) & 32 (23, 49) & 5600 & 71 (59, 94) & 0.85 & 26 \\
     & mbt2018 & 5 & 31.73 & SSIM & no & 0.94 (0.82, 1.0) & 32 (23, 46) & 8154 & 55 (-231, 89) & 1.23 & 27 \\
     & mbt2018 & 5 & 31.73 & SSIM & yes & 0.94 (0.83, 1.0) & 32 (23, 44) & 7415 & 59 (24, 94) & 1.12 & 26 \\
     & mbt2018 & 6 & 0.0483 & MSE & no & 0.93 (0.84, 1.0) & 33 (25, 50) & 8838 & 55 (-294, 93) & 1.33 & 27 \\
     & mbt2018 & 6 & 0.0483 & MSE & yes & 0.93 (0.84, 1.0) & 34 (25, 50) & 8251 & 58 (47, 93) & 1.25 & 28 \\
     & mbt2018 & 6 & 60.5 & SSIM & no & \textbf{0.96 (0.89, 1.0)} & 33 (22, 47) & 11435 & \textbf{35 (-145, 69)} & 1.73 & 27 \\
     & mbt2018 & 6 & 60.5 & SSIM & yes & \textbf{0.96 (0.89, 1.0)} & 33 (25, 47) & 11090 & \textbf{38 (-270, 73)} & 1.67 & 27 \\
     & mbt2018 & 7 & 0.0932 & MSE & no & 0.95 (0.85, 1.0) & 35 (24, 51) & 12441 & 35 (-114, 91) & 1.88 & 26 \\
     & mbt2018 & 7 & 0.0932 & MSE & yes & \textbf{0.95 (0.88, 1.0)} & 35 (26, 52) & 11626 & \textbf{39 (16, 83)} & 1.76 & 27 \\
     & Elic2022 &  & 0.03 & MSE & yes & \textcolor{blue}{\textbf{0.92 (0.78, 1.0)}} & 33 (23, 50) & 5639 & \textcolor{blue}{\textbf{71 (64, 94)}} & 0.85 & 115 \\
     & Elic2022 &  & 0.045 & MSE & yes & 0.93 (0.81, 1.0) & 34 (24, 51) & 6934 & 64 (57, 95) & 1.05 & 116 \\
     & Elic2022 &  & 60.5 & SSIM & yes & 0.95 (0.84, 1.0) & 32 (23, 46) & 9154 & 48 (4, 80) & 1.38 & 112 \\
  
    \bottomrule
    \end{tabular}
    \end{adjustbox}
    \caption{Image quality metrics (SSIM and PSNR), compression rate and decompression times for JPEG, JPEG-2000, JPEG-XL, and the deep learning models on the balanced dataset of tissue patches. \hl{The deep learning models are either used out of the box from CompressAI, Pre = 1 (pretrained), or trained on pathology data, Pre = 0. The models trained on pathology data are trained with and without augmentation, at different quality levels, and with different distortion losses. }All values are the average values for a patch in the balanced dataset, except the \textit{total kB} which represents the total size of all compressed patches in kB, the decompression time which is the average for all patches after the first 10 and the minimum and maximum SSIM and PSNR which are values for a single patch. The bmshj2018-factorized, bmshj2018-hyperprior, and mbt2018-mean \hl{models'} quality level\hl{s} (q) \hl{are} between 5-7. \hl{The size reduction is }compared to JPEG\hl{-compression}. The top performing deep learning models are marked in \textbf{bold}, while the models that save the most space compared to JPEG are marked in \textcolor{blue}{\textbf{blue}}. \hl{The best performing deep learning models marked in \textbf{bold} had SSIM values on par with JPEG, while saving on average $\sim$35-40\% space per patch compared to JPEG. However, the models had large ranges between the patches that saved the most and least space, and the decompression times of the deep learning-based models were slower than JPEG, JPEG-2000, and JPEG-XL. }Abbreviations: MSE = mean squared error, SSIM = structural similarity index, PSNR = peak signal-to-noise ratio, Pre = pretrained, Q = quality level, bpp = bits per pixel, D = distortion, aug = augmentation, Arch = architecture, min = minimum, max = maximum, dec = decompression, factorized = bmshj2018-factorized, hyperprior = bmshj2018-hyperprior, mbt2018 = mbt2018-mean, Elic2022 = Elic2022Chandelier.}
\label{tab:results_tissueCompt}
\end{table*}

\subsubsection{Experiments}
\begin{figure}[H]
    \centering
    \includegraphics[width=0.5\linewidth]{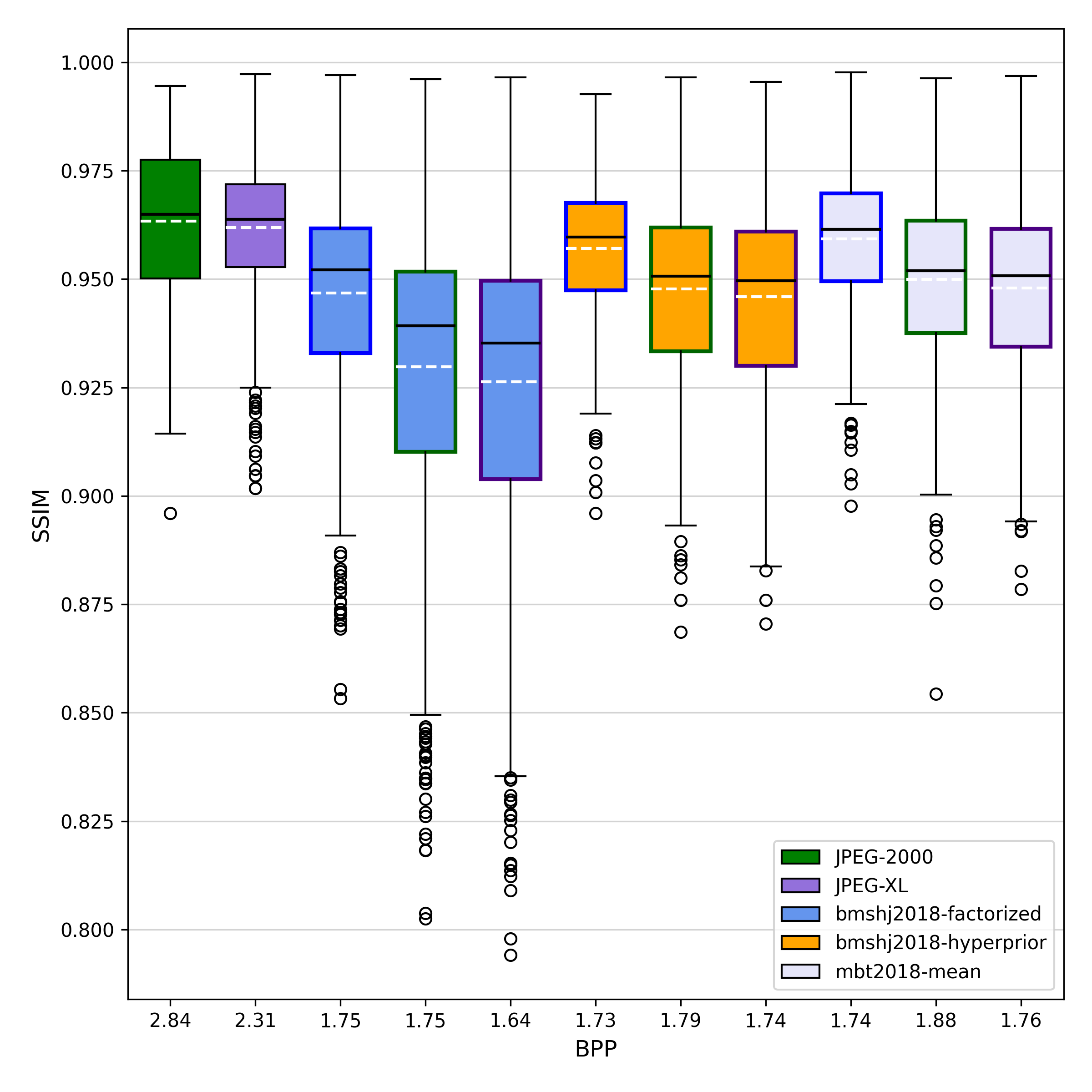}
    \caption{Structural similarity index (SSIM), and average bits per pixel (BPP) per patch, in the balanced test set with JPEG-2000, JPEG-XL, and the bmshj2018-factorized (blue boxes), bmshj2018-hyperprior (orange boxes), and mbt2018-mean (light purple boxes) models optimized for MSE in the rate-distortion loss, at quality level 7/lambda value 0.0932. Outlines in blue represent pretrained models, outlines in green represent models trained without augmentation, and the boxes outlined in dark purple represent the models trained with augmentation. The full black line represents the median, and the stapled white line represents the mean.}
    \label{fig:boxplot_tissueComp}
\end{figure}

\noindent\textbf{Value of training on pathology data}:
The pretrained models and the corresponding models trained on pathology data achieved similar average SSIMs and PNSRs on the balanced test set, but the pretrained models achieved slightly higher average SSIMs and had a higher minimal SSIM than the trained models (Fig. \ref{fig:boxplot_tissueComp}). \\
\\

\begin{figure*}
    \centering
    \includegraphics[width=0.8\linewidth]{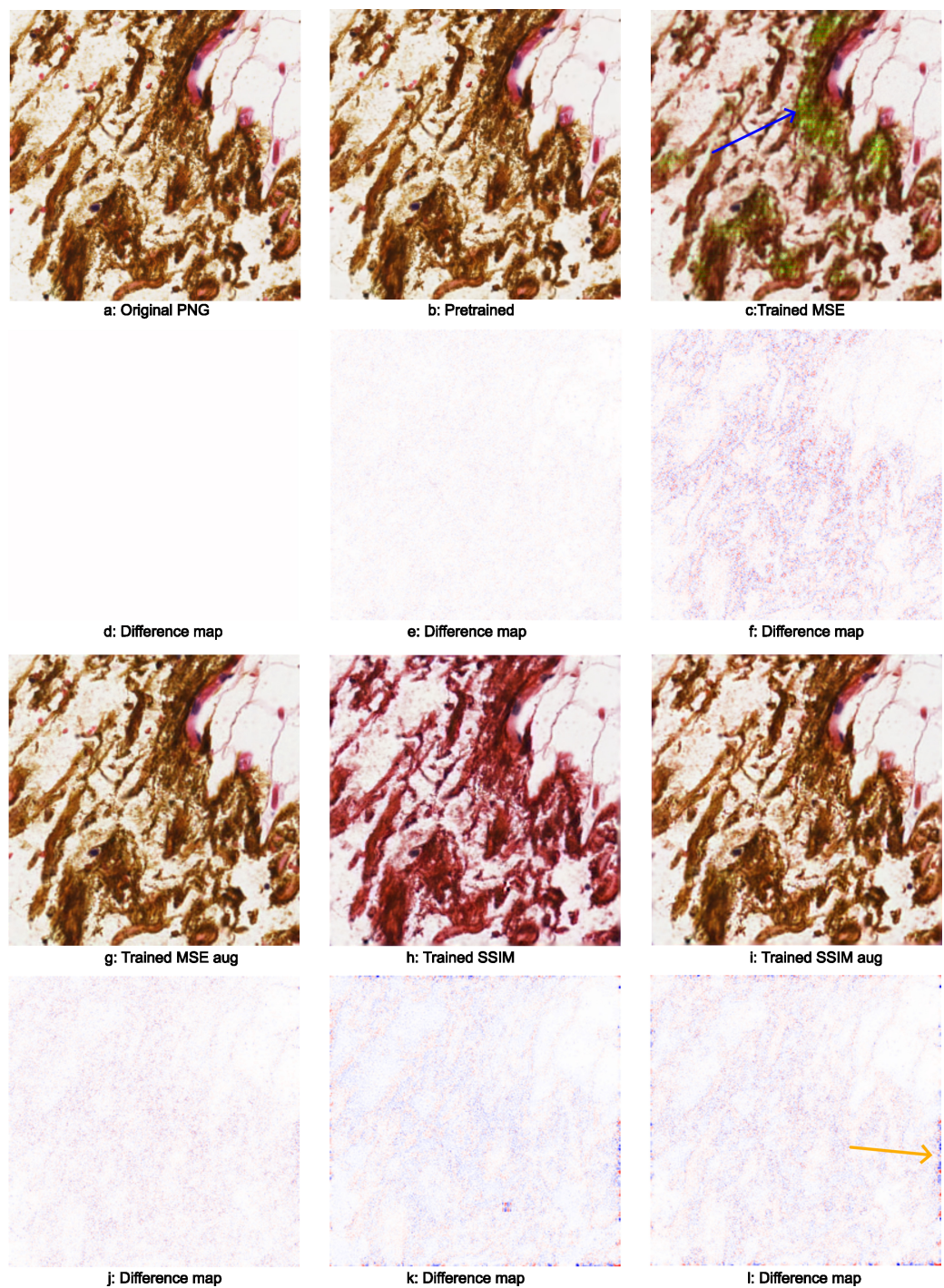}
    \caption{A patch where the trained models performed poorly. The patch is from an HE-stained slide, but is probably extracted from a region with an artifact (pen marks), and does thus not display regular hematoxylin nor eosin colors. a: The original PNG patch. b: The patch compressed with the pretrained mbt2018-mean model at quality level 7 (SSIM 0.95). c: The patch compressed with the mbt2018-mean model trained on pathology data with MSE as the distortion loss at quality level 7 (SSIM 0.85). d: The difference map between the original PNG patch in a and itself. e: The difference map between the original PNG patch in a and the patch in b. f: The difference map between the original PNG patch in a and the patch in c. g: The patch compressed with the mbt2018-mean model trained on pathology data with MSE as the distortion loss and trained with augmentation (SSIM 0.94). h: The patch compressed with the mbt2018-mean model trained on histopathology data with SSIM as the distortion loss (SSIM 0.91). i: The patch compressed with the mbt2018-mean model trained on histopathology data with SSIM as the distortion loss and trained with augmentation (SSIM 0.93). j: The difference map between the original PNG patch in a and the patch in g. k: The difference map between the original PNG patch in a and the patch in h. l: The difference map between the original PNG patch in a and the patch in i. 
    Row 2 and 4 show the difference map between the decompressed patch and the original patch in gray scale (blue = negative pixel intensity in difference map, red = positive pixel intensity and white is unchanged). The blue arrow in the top right patch points to a color artifact, and the yellow arrow in the bottom right points to edge artifacts seen in patches compressed and decompressed with models optimized for SSIM. Abbreviations: SSIM = structural similarity index, MSE = mean squared error. The patch in this illustration is from the Her2Neg\_Case\_22 in the HER2 tumor ROIs dataset~\cite{Farahmand22}, with a \href{https://creativecommons.org/licenses/by/4.0/}{CC BY 4.0} license, available at The Cancer Imaging Archive (TCIA)\cite{clark2013cancer}, and has been modified for this illustration }
    \label{fig:worst_patch}
\end{figure*}

\noindent\textbf{The effect of the distortion loss}:
The models trained with SSIM as the distortion loss achieved slightly higher average SSIMs, but lower PSNRs, than the models trained with MSE as the distortion loss on the balanced test data set, for similar bpps (Table \ref{tab:results_tissueCompt}). However, edge artifacts were observed with the models optimized with SSIM in the distortion loss as shown in Fig.~\ref{fig:worst_patch}.\\

\noindent\textbf{The effect of augmentation}:
The models trained with and without augmentation achieved similar average SSIM scores. The minimum SSIM for the bmshj2018-hyperprior and mbt2018-mean models optimized for MSE with lambda 0.0932 increased when augmentation was added. The mbt2018-mean minimum SSIM increased from 0.85 to 0.88 when augmentation was added (Fig. \ref{fig:boxplot_tissueComp}). The large color artifacts seen for two of the models on the patch in Fig. \ref{fig:worst_patch}, were removed when augmentation was added. The average compressed size of patches compressed with the models with augmentation were slightly lower than with the models trained without augmentation (Table \ref{tab:results_tissueCompt}).\\
\\
\noindent\textbf{Influence of magnification}:
The bmshj2018-factorized, bmshj2018-hyperprior and mbt2018-mean models optimized for MSE at quality level 7 were evaluated on the magnification test sets, and a difference in SSIM scores between the magnifications was observed (Table \ref{tab:tissue_comp_mag}). 
The SSIM scores for the pretrained models at quality level 7 and the corresponding trained models (without augmentation and optimized for MSE) increased with higher magnification. The pretrained models achieved higher SSIMs than the trained models at magnification $\times$1.25, $\times$2.5, $\times$5, $\times$10, and $\times$40, while the trained models achieved the same, or higher SSIMs than the pretrained models, at magnification $\times$20 (Table \ref{tab:tissue_comp_mag}).

A difference in average saved space per patch compared to JPEG was also observed between the magnifications. For the pretrained mbt2018-mean model, the lowest average saved space was observed at $\times$2.5 where it was 34\%, and the highest at $\times$20, where it was 49\% (Table \ref{tab:tissue_comp_mag}).\\
\\

\begin{table}[h!]
    \centering
    \begin{tabular}{lllllllll}
        \toprule
        Arch. & Metric & Pre. & $\times$1.25 & $\times$2.5 & $\times$5 & $\times$10 & $\times$20 & $\times$40 \\
        \midrule
        Fac & SSIM &  Y & 0.92 & 0.93 & 0.95 & 0.96 & 0.96 & 0.96 \\
                   & Space & Y & 39 & 34 & 36 & 38 & 34 & 46 \\
                   & SSIM & N & 0.90 & 0.90 & 0.93 & 0.94 & 0.96 & 0.95 \\
                   & Space & N & 39 & 36 & 31 & 29 & 17 & 47 \\
        \midrule
        Hyp & SSIM & Y & 0.94 & 0.95 & 0.96 & 0.97 & 0.96 & 0.96 \\
                   & Space & Y & 39 & 35 & 37 & 43 & 47 & 46 \\
                   & SSIM & N & 0.93 & 0.93 & 0.95 & 0.96 & 0.97 & 0.95 \\
                   & Space& N & 37 & 33 & 32 & 35 & 41 & 41 \\
        \midrule
        Mbt & SSIM & Y & 0.94 & 0.95 & 0.96 & 0.97 & 0.96 & 0.97 \\
                & Space & Y & 39 & 34 & 38 & 43 & 49 & 46 \\
                & SSIM & N & 0.94 & 0.94 & 0.95 & 0.96 & 0.97 & 0.95 \\
                & Space & N & 35 & 30 & 30 & 34 & 42 & 36 \\
        \bottomrule
    \end{tabular}
    \caption{Average structural similarity indices (SSIM) and saved space for the pretrained models (Pre.) and corresponding models trained on pathology data optimized for mean squared error at quality level 7, on the magnification test sets. The SSIM and saved space changes with magnification level. Abbreviations: Arch. = architecture. Pre. = pretrained. Y = yes. N = no. Fac = bmshj2018-factorized. Hyp = bmshj2018-hyperprior. Mbt = mbt2018-mean. The Space metric is the average saved space per patch compared to JPEG (\%).}
    \label{tab:tissue_comp_mag}
\end{table}

\noindent\textbf{Comparison with non-learning methods}:
On the balanced test set, JPEG-2000 and JPEG-XL had the highest average PSNR, and average SSIM, though many models had very similar average SSIM. However, they also had a higher bpp, and thus saved less space than the top performing deep learning methods on this test set (Fig. \ref{fig:boxplot_tissueComp} and Table \ref{tab:resultsTissueSeg}). JPEG-XL and JPEG-2000 achieved similar image quality metrics, but JPEG-XL reached a lower average bpp than JPEG-2000.\\
\\
\noindent\textbf{Worst case - artifacts}:
In Fig. \ref{fig:worst_patch}, we present the patch with the lowest SSIM for the mbt2018-mean model optimized for MSE with lambda 0.0932, and trained without augmentation from the balanced dataset. The figure shows the patch compressed and decompressed with the pretrained version of this model and with the trained version of this model with SSIM and MSE distortion loss and trained with, and without augmentation. On this patch, the two presented models trained without augmentation performed poorly, with prominent color artifacts. The same patch compressed with JPEG, JPEG-2000 and JPEG-XL all achieved visually pleasing results, and SSIMs of 0.93, 0.98 and 0.93 for JPEG, JPEG-2000 and JPEG-XL, respectively.

\begin{figure*}
    \centering
    \includegraphics[width=0.9\linewidth]{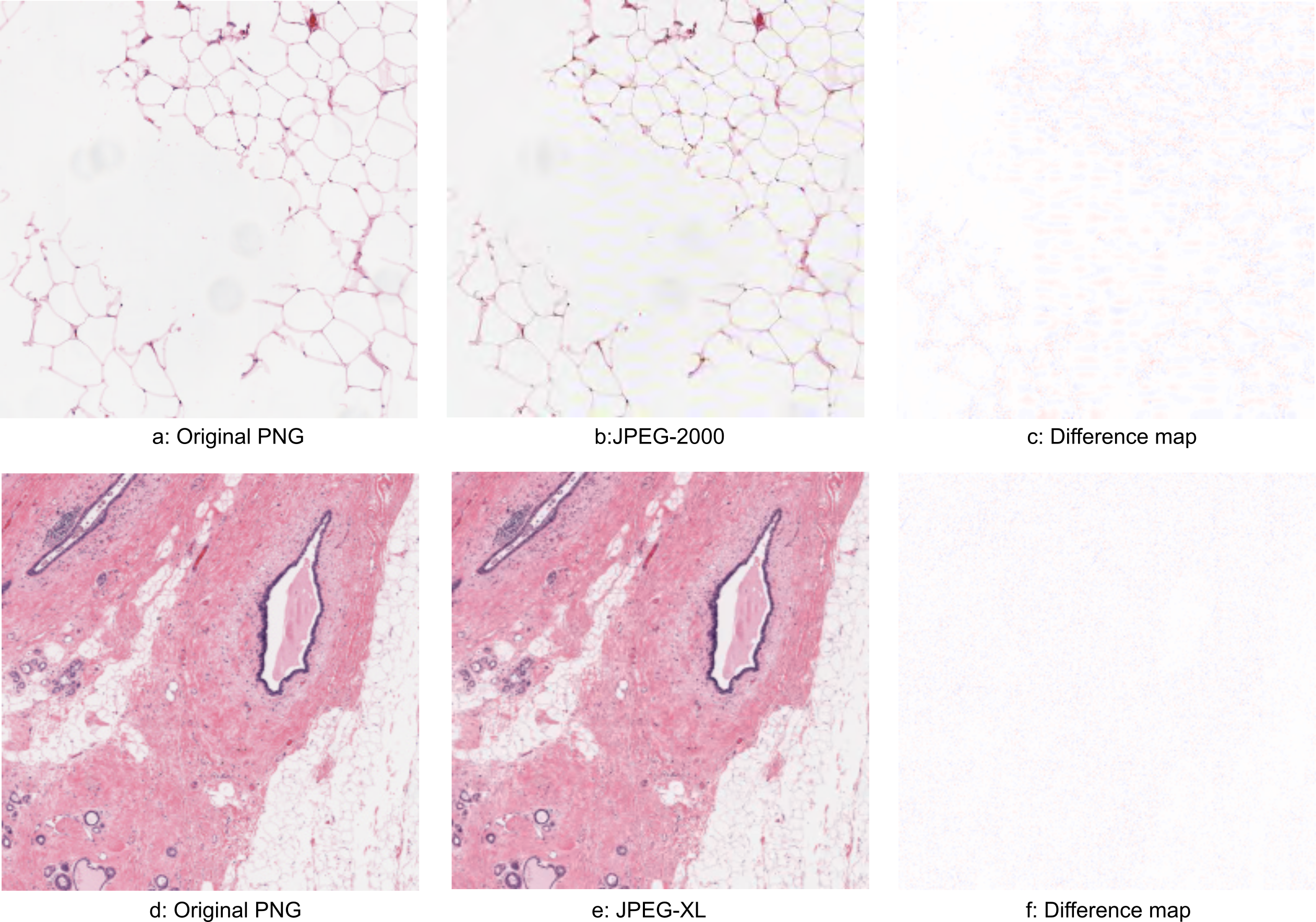}
    \caption{The JPEG-2000 and JPEG-XL patches with the lowest structural similarity index (SSIM) in the balanced dataset. a: Original PNG. b: Patch compressed with JPEG-2000 (SSIM 0.90). c: Difference map between original PNG in a and patch in b. d: Original PNG. e: Patch compressed with JPEG-XL (SSIM 0.90). f: Difference map between original PNG in d and patch in e. No large artifacts are present in either b nor e. The JPEG-XL is visually clearer than the JPEG-2000 patch compared to their original PNGs. The patches in this illustration are from the Her2Neg\_Case\_22 in the HER2 tumor ROIs dataset~\cite{Farahmand22}, which is available at The Cancer Imaging Archive (TCIA)\cite{clark2013cancer}, with a \href{https://creativecommons.org/licenses/by/4.0/}{CC BY 4.0} license,  and have been modified for this illustration }
    \label{fig:worstJP2JXL}
\end{figure*}

Neither the worst JPEG-2000 patch, nor the worst JPEG-XL patch in the balanced dataset had any large artifacts (Fig. \ref{fig:worstJP2JXL}).

\subsubsection{Runtime}
The average decompression time per patch \hl{was} below 30 ms in the balanced test set for all the bmshj2018-factorized, bmshj2018-hyperprior and mbt2018-mean models, whereas the decompression time was much slower for the Elic2022Chandelier models (Table \ref{tab:results_tissueCompt}). Decompression was done with CompressAI's model.decompress function~\cite{begaint2020compressai}. JPEG and JPEG-XL had much faster decompression times, on average 2 and 3 ms, respectively. The experiments were done on a Quadro RTX 6000 GPU and Intel Xeon Gold 6230 CPU @ 2.10 GHz.

\subsection{Glass removal and tissue compression}
Replacing glass with white pixels saved space for JPEG and JPEG-XL. Replacing glass with empty tiles led to a further size reduction for JPEG, but not for JPEG-XL, thus JPEG-XL compresses single color images better than JPEG. Compression with the pretrained mbt2018-mean model at quality level 7 on \textit{patch pyramids}, resulted in a size reduction of 43-80\% compared to the JPEG-compressed image pyramids with intact glass (Fig. \ref{fig:glassRemovalBoxplot}). The size reduction was similar for the slides with intact glass (43-72\%), the slides with glass replaced by white pixels (44-75\%), and the image slides where glass tiles were removed (45-80\%) for this pretrained model (Fig. \ref{fig:glassRemovalBoxplot}). For the latter case, the total size of all 21 WSIs was 2.8 GB. This represents a total size reduction of 57\% compared to the total size of all 21 JPEG-compressed pyramids with intact glass.

\clearpage

\section{Discussion}
This study investigated modern image compression techniques, including deep learning, for reducing the file size of WSIs. The large size of WSIs is a considerable and increasing challenge in pathology laboratories worldwide. 
\subsection{Tissue segmentation}
The tissue segmentation model was robust on all magnification levels, and achieved a dice score of 0.980$\pm$0.034 at magnification $\times$2.5. Compared to the model developed by Jurgas \textit{et al.}~\cite{jurgas2023robust}, the proposed model achieved higher dice scores, but was only tested on HE-stained slides. Its performance was similar to the results reported by Riasatian \textit{et al.}~\cite{riasatian2020comparative}. Still, different datasets were used and the results are therefore not directly comparable.

The magnification level influenced the tissue segmentation mask as shown in Fig.~\ref{fig:her2neg_case_22}. At low magnifications, most tissue areas were identified, resulting in a high recall but low precision. Segmentations at high magnifications had more accurate borders, and consequently higher precisions. However, some tissue structures, for example fat and small tissue fragments, were often missed\hl{, leading to a higher false negative rate.} This recall/precision trade off between magnifications may depend on the tissue structures included in a patch during inference. Depending on the magnification, a patch can include very few cells, or include large structures. High magnification patches containing adipose tissue may represent only a few cell walls, or only the inside of fat cells. The neural network can consequently segment it incorrectly as glass. 
The magnification level also influence\hl{d} the runtime of the tissue segmentation. The runtime was lowest at magnification $\times$1.25 and highest at $\times$20, when testing at $\times$1.25, $\times$2.5, $\times$5, $\times$10, and $\times$20. Thus, to optimize runtime\hl{, and limit tissue loss, especially adipose tissue loss}, a lower magnification \hl{is likely} prefer\hl{red}. 

The tissue segmentation model achieved high dice scores, but upon visual inspection, some limitations were observed. Even though the dataset included WSIs with artifacts, and the use of artifact-mimicking data augmentation, some artifacts were still segmented as tissue by the model. In addition, for some WSIs, pixels at the tissue borders were not segmented. A morphological post processing stage, such as dilation and erosion, could have helped to reduce these mistakes. 

Brightness variations affected the neural network-based segmentations. However, the neural network segmentation model was more robust to brightness variations than regular color thresholding. Color thresholded tissue segmentation created precise segmentations at some brightness levels, but failed at others. The brightness may differ between WSIs, making it difficult to set a color threshold value that works for all WSIs in a dataset. It also often excludes fat, and lightly colored stroma, which was observed in the test WSI in Fig. \ref{fig:brightness}. 
A neural network can be trained to be much more robust in terms of brightness changes and other variations as the experiment in Fig. \ref{fig:brightness} showed. \hl{It would also be possible to train a tissue segmentation model for multiple stains, e.g. immunohistochemistry, whereas a  non-trainable method could potentially be more difficult to create for multiple stains.}
However, the \hl{results of the }deep learning segmentation model \hl{are dependent on the magnification level, and it }also struggled with the test WSI at brightness factor -100. Robustness towards brightness is important, as variations in stain intensity and differences between scanners may influence the brightness of an image. If more aggressive color augmentation was added during training, this could potentially be fixed.

The deep learning models were trained on manual annotations. Manual annotations are error prone and time-consuming to make exact, even with the use of active learning. In addition, WSIs may include tissue with complex and irregular structures, making them time-consuming and difficult to annotate. This could have influenced the tissue segmentation results, both in terms of the qualitative results, and the quantitative as the ground truth may not always be completely accurate.

In this work, tissue segmentation was done for glass removal and consequently reduction of WSI file size. \hl{It plays a crucial role, as it determines which information is permanently removed from the WSI, potentially resulting in undesirable tissue loss.} \hl{T}issue segmentation is also a very important pre-processing step in most AI analysis of WSIs. It is often used to separate tissue from glass, to ensure that\hl{ }neural network training is only done on patches containing tissue. 

\subsection{Glass removal}
For all WSIs, the highest file size reduction was achieved with replacement of all-glass tiles with empty tiles. However, there was no large difference in file size reduction between the two glass removal methods for JPEG-XL compression. This indicates that JPEG-XL compresses single color tiles effectively. When compressing a $512 \times 512$ white patch with JPEG and JPEG-XL, the JPEG-XL-compressed patch consumed less than 200 bytes, whereas the JPEG-compressed patch consumed more than 4 kB. 

The file size reduction by glass removal was highly dependent on the WSI and the amount of glass in the image, regardless of glass replacement strategy and compression format (JPEG or JPEG-XL).
For the 21 WSIs in the dataset, the size reduction varied from 0.3 to 71\%.
Varying size reduction may be due to complex tissue structures within the slides. Slides with spatially dispersed tissue areas are very likely to produce tiles containing small amounts of tissue and thus cannot be replaced by empty tiles. The evenness of the glass areas in the SVS files also influenced the size reduction, as some of the original SVS files had a very even glass color, whereas others did not. Replacing very even glass background with white tiles does not lead to a large size reduction. The compression format (JPEG or JPEG-2000) of the original SVS files may also have influenced the size reduction. 

\subsection{Tissue compression}
The models with the highest image quality metrics saved less storage space than the models with low image quality metrics. 
The best performing deep learning models on the balanced test set achieved similar average SSIM and PSNR values to JPEG-XL and JPEG-2000. 
The mbt2018-mean pretrained compression model at quality level seven achieved on average 42\% saved space on the tissue patches in the balanced test set, compared to JPEG, while keeping the average SSIM and PSNR at 0.96 and 36, respectively.
This is more than double the file size reduction compared to JPEG-2000 and JPEG-XL on the same data (14\% and 17\%). The total size on disk for all tissue patches was 18 678 kB, 18 812 kB, 15 293 kB, and 11 510 kB, with JPEG, JPEG-2000, JPEG-XL, and mbt2018-mean pretrained, respectively. \hl{The mbt2018-mean pretrained compression model at quality level 5 and 6 also outperformed JPEG and JPEG-XL at quality 70 and 80 on the patched dataset (Supplementary material Figure 1-2)}.
This demonstrates the potential of deep learning-based image compression.

However, it was observed that the deep learning models can produce image artifacts, as shown in \hl{Fig.} \ref{fig:worst_patch}.
The brown colors in this patch are usually not present in HE-stained slides, and may contribute to the WSI trained deep learning models' poor performance. The models trained with augmentation did not present with the same color artifact for this patch, \hl{indicating the importance of augmentation,} but a full qualitative evaluation of all patches in the test set would have to be performed to exclude the possibility of large color artifacts for the models trained with augmentation.
Edge artifacts were also observed with models optimized for SSIM, which may be due to the way SSIM is calculated at the image borders. The SSIM score for the patch with the large color artifact in Fig. \ref{fig:worst_patch} was quite high despite color artifacts and edge artifacts, indicating that SSIM is not able to properly detect these. \hl{The SSIM compares the similarity between two images, but it does not necessarily give perceptually meaningful information corresponding to diagnostic information.}

The compression results on the balanced test set were lower than those presented by Fisher \textit{et al.}, who reported a file size reduction of 76\% compared to JPEG with quality factor 80, with a MS-SSIM equal to 0.99. We also achieved high MS-SSIM values for models at low quality levels and low lambda values. For example, for the pretrained bmshj2018-factorized model at quality level 5, we achieved an MS-SSIM of 0.98, and saved 71\% compared to JPEG. Differences in datasets and magnification level could, however, have contributed to the different results, and models may perform differently at different bpps. For a proper comparison, the model of Fischer \textit{et al.} would have had to be tested on our dataset and with the same metrics, compared to the same JPEG quality, and ideally at many bpps.

JPEG-2000 and JPEG-XL achieved higher average SSIM and PSNR than JPEG. JPEG-2000 and JPEG-XL also saved space compared to JPEG when measuring the average saved space for a patch. However, the total size of the balanced dataset was higher with JPEG-2000 than with JPEG-compression. The file size of some patches increased by 74\%, while some decreased by 92\%, indicating a large difference in compression efficiency between patches for JPEG-2000. Some patches increased in size compared to JPEG for JPEG-XL, but the increase was much smaller (3\%). Compared to the deep learning models, JPEG-XL achieved similar average SSIM and PSNR scores to the best deep learning models. The minimum SSIM score was, however, higher for JPEG-XL than the best deep learning models on the balanced test set, indicating that it may be more robust. The poorest resulting patches in terms of SSIM with JPEG-XL and JPEG-2000 (Fig. \ref{fig:worstJP2JXL}) indicate that these compression methods perform well in terms of image quality even at their worst.

Some loss in image quality may be inevitable, which is also true for today's lossy JPEG compression. To determine how much loss could be accepted in the clinic, a qualitative study may be needed. 
The use cases for the compressed images will also affect the rate distortion trade off of the compression models. 
Efficient training of new machine learning models may favor a high distortion and high compression rate if a perfect reconstructed image is not needed for the downstream task. In the clinic, however, low distortion is necessary, as even small artifacts may influence a pathologist's evaluation of a WSI. In addition to rate and distortion, decompression time would be important to consider in many use cases. In image compression, the trade off between image quality, compressed size, and compression and decompression time needs to be optimized for the given task. 

The tested compression methods in CompressAI had much slower decompression than the CPU-based JPEG and JPEG-XL implementations. \hl{This could be due to differences in the transform algorithms; JPEG and JPEG-XL are based on the discrete cosine transform whereas the deep learning models are convolutional neural networks which perform many arithmetic operations and thus require a GPU to run quick enough for practical use. }\hl{In addition},\hl{ }CompressAI \hl{ and PyTorch are research tools, and }might not be fully optimized for runtime performance \hl{whereas the JPEG and JPEG-XL libraries are optimized for runtime performance and production use}.

\subsection{Glass removal and tissue compression}
Fig. \ref{fig:glassRemovalBoxplot} shows that glass removal was very beneficial for JPEG and JPEG-XL compression, whereas the benefit of glass removal was very small when using the pretrained mbt2018-mean AI model at quality level seven. This indicates that this AI model can compress glass tiles efficiently.

The \textit{patch pyramids} contain no metadata, whereas the image pyramids do. This may have led to a slightly higher size reduction for the \textit{patch pyramids}. On the other hand, patches at the border were padded, which may have increased the amount of data. 

\subsection{Strengths and weaknesses}
In this study,\hl{ } \hl{mainly} HE-stained \hl{SVS} slides were \hl{included}. HE is the routine staining in diagnostic pathology, but ideally a compression method should also work on other staining types, such as immunohistochemistry. \hl{This would require a dataset including slides with different stains when training and evaluating models, and should be explored in future studies. Such a dataset could potentially make the models more robust to naturally occurring stain artifacts, like the brown color observed in Fig.} \ref{fig:worst_patch}. \hl{One breast cancer CK 5/6-stained slide was evaluated with the tissue segmentation model at $\times$2.5 (Supplementary material, Fig. 3). The model performed satisfactory on epithelial and stromal regions, and also included foci with abundant inflammatory cells, but incorrectly excluded large areas of adipose tissue. A \textit{patch pyramid} of the slide with glass was also compressed with the pretrained mbt2018-mean model at quality factor 7. It saved 56\% space compared to the corresponding JPEG-compressed WSI. This is similar to the results on the HE-stained slides in this paper. The average SSIM on a balanced test set of 672 tissue patches with this model was 0.96, which is on par with the average SSIM for the balanced HE-stained test set, indicating that the CompressAI models trained on natural images work equally well on non-HE stains. The methods are described in the Supplementary material.}
The dataset included 21 WSIs from seven different organs. This may be sufficient to make a good tissue segmentation model. However, more data may be needed to cover different tissue morphologies and artifacts found in WSIs. 
The WSIs in this study were already compressed, resulting in a double compression in the created image pyramids and compressed patches. Ideally, one should use uncompressed images as input, and thus only compress the images once. However, this is challenging due to the extreme size and unavailability of uncompressed WSIs.

A comprehensive model and JPEG evaluation was performed. Many deep learning models were evaluated at multiple quality levels. Artifacts were investigated, and a WSI specific artifact augmentation for WSIs were added during training of the tissue segmentation model. Glass replacement with both white pixels and empty tiles were evaluated. All experiments were performed on open data, and the code and tools are openly available. 

\subsection{Future work}
This study strengthens our hypothesis of the possibility of reducing WSI file sizes using deep learning and modern image compression while maintaining high image quality. However, before clinical implementation of such compression models, more work is needed.

For tissue segmentation and glass removal, a larger \hl{dataset}, potentially also including other staining methods than HE may be needed.
More robust training with heavier augmentations may also increase the robustness of the tissue segmentation model.
Segmenting fat tissue as a separate class may be beneficial for other tasks than glass removal, such as pre-processing for further AI analysis.

In this study, image compression using deep learning has shown promise. However, since deep learning compression models will never be guaranteed artifact-free, further work should focus on detection and an automatic fallback strategy if the image quality of the decompressed image is too poor.

To facilitate the use of these models, at least for digital pathology research, WSI software, such as OpenSlide~\cite{GoodeAdam2013OAvs}, QuPath~\cite{BankheadPeter2017QOss}, and FAST~\cite{SmistadErik2015Fffh, SmistadErik2019HPNN}, need to support these neural network compression models.
Storing neural network compressed tiled image pyramids can be achieved using the TIFF file format. 

\section{Conclusion}
Glass removal with deep learning tissue segmentation reduces the file size considerably for JPEG and JPEG-XL-compressed WSIs. JPEG-XL achieved much higher compression rates on the WSIs than JPEG. 
Deep learning-based image compression shows promise\hl{ }and could possibly save considerably more space than traditional JPEG and JPEG-XL. However, neural network\hl{-}based compression would have to be supported in WSI software libraries, and the decompression times are currently much higher, even with a GPU. Before clinical implementation, thorough qualitative studies need to be performed, and a solution for fallback in case of artifacts needs to be included. 

\section{Data availability statement}
Results in this work were derived from the HER2 tumor ROIs (DOI: 10.7937/E65C-AM96), CPTAC-COAD (DOI: 10.7937/TCIA.YZWQ-ZZ63), CMB-CRC (DOI: 10.7937/DJG7-GZ87), CMB-LCA (DOI: 10.7937/3CX3-S132), CPTAC-OV (DOI: 10.7937/TCIA.ZS4A-JD58), CPTAC-PDA (DOI: 10.7937/K9/TCIA.2018.SC20FO18), CMB-PCA (DOI: 10.7937/25T7-6Y12), CPTAC-UCEC (DOI: 10.7937/K9/TCIA.2018.3R3JUISW), \hl{HE-vs-MPM (DOI: 10.7937/3FYC-AC78)} datasets in The Cancer Imaging Archive.

\section{Acknowledgments}
Data used in this publication were generated by the National Cancer Institute Clinical Proteomic Tumor Analysis Consortium (CPTAC).
Data used in this publication were generated by the National Cancer Institute’s Cancer Moonshot Biobank.

\section{Funding}
The work was funded by The Liaison Committee for Education, Research, and Innovation in Central Norway [Grant Number 2020/39645], the Joint Research Committee between St. Olavs hospital and the Faculty of Medicine and Health Sciences, NTNU (FFU) [Grant Number 2021/51833], and the Clinic of Laboratory Medicine, St. Olavs hospital, Trondheim University Hospital, Trondheim, Norway. 

\printbibliography
\end{document}


\pagestyle{fancy}
\fancyhf{}
\fancyhead[L]{This paper is currently under consideration in a scientific journal.}
\date{}  
\author[]{}

\maketitle
\thispagestyle{fancy}

\section{Dataset}
\hl{The images used in the article \textit{Deep learning-based compression of giga-resolution whole slide images} are described in Table} \ref{tab:dataset}. \hl{The train and test set splits for the tissue compression is described in Table} \ref{tab:compression_dataset}\hl{, and the magnification and balanced test sets for the tissue compression is described in Table }\ref{tab:test_data}\hl{. The number of patches at different magnifications for the tissue segmentation dataset is described in Table }\ref{tab:patchnumber}.

\section{Rate-distortion curve}
\begin{figure*}[h!]
    \centering
    \includegraphics[width=0.5\linewidth]{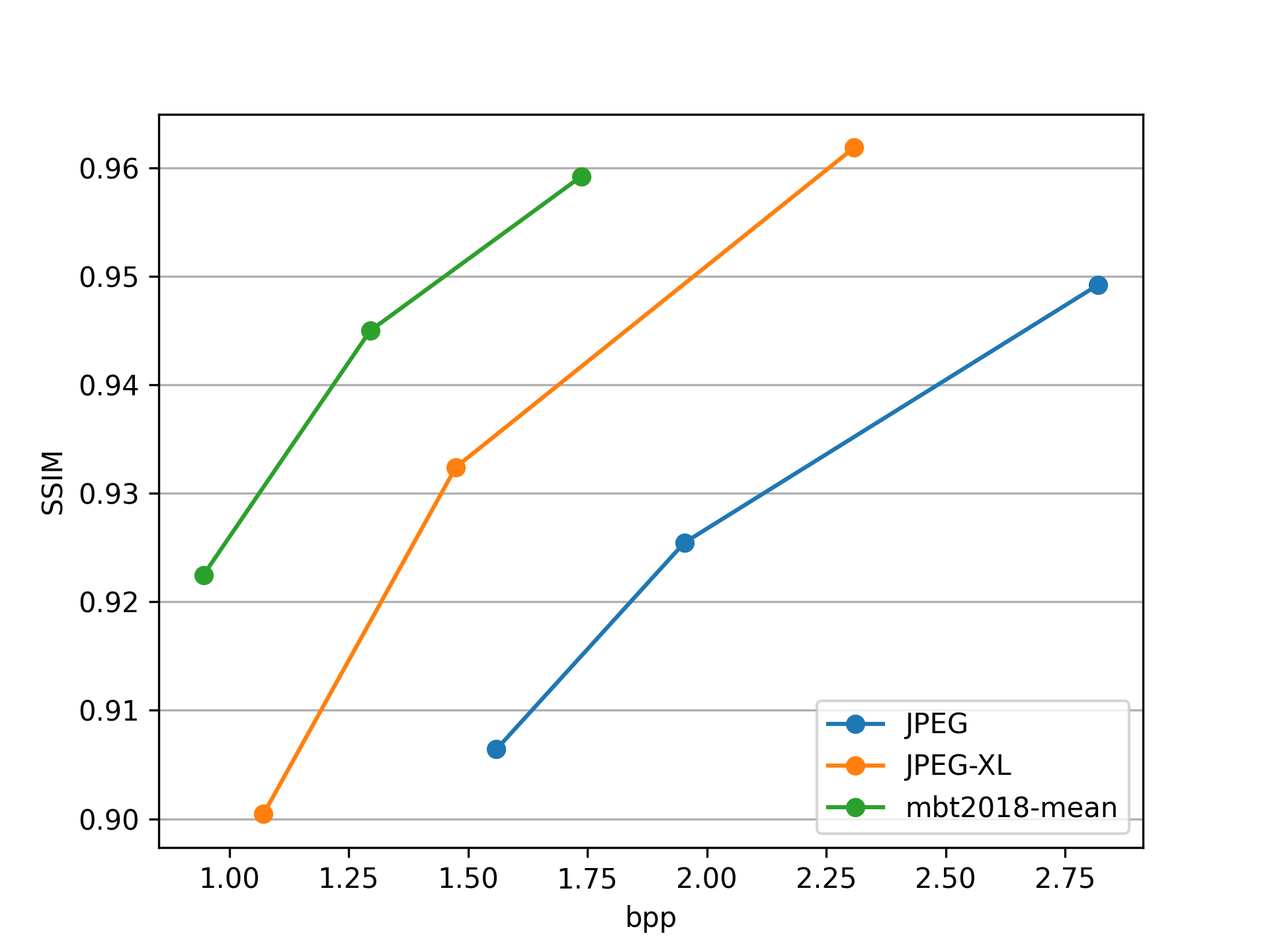}
    \caption{\hl{Rate distortion curve for JPEG, JPEG-XL and the pretrained model mbt2018-mean q7 on the balanced patch dataset. SSIM against bpp. The points for JPEG and JPEG-XL are at quality factors 70, 80, and 90, whereas the deep learning-based model is at quality level 5, 6, and 7. Abbreviations: SSIM = structural similarity index, bpp = bits per pixel.}}
    \label{fig:supmat_bppSSIM}
\end{figure*}

\begin{figure*}[h!]
    \centering
    \includegraphics[width=0.5\linewidth]{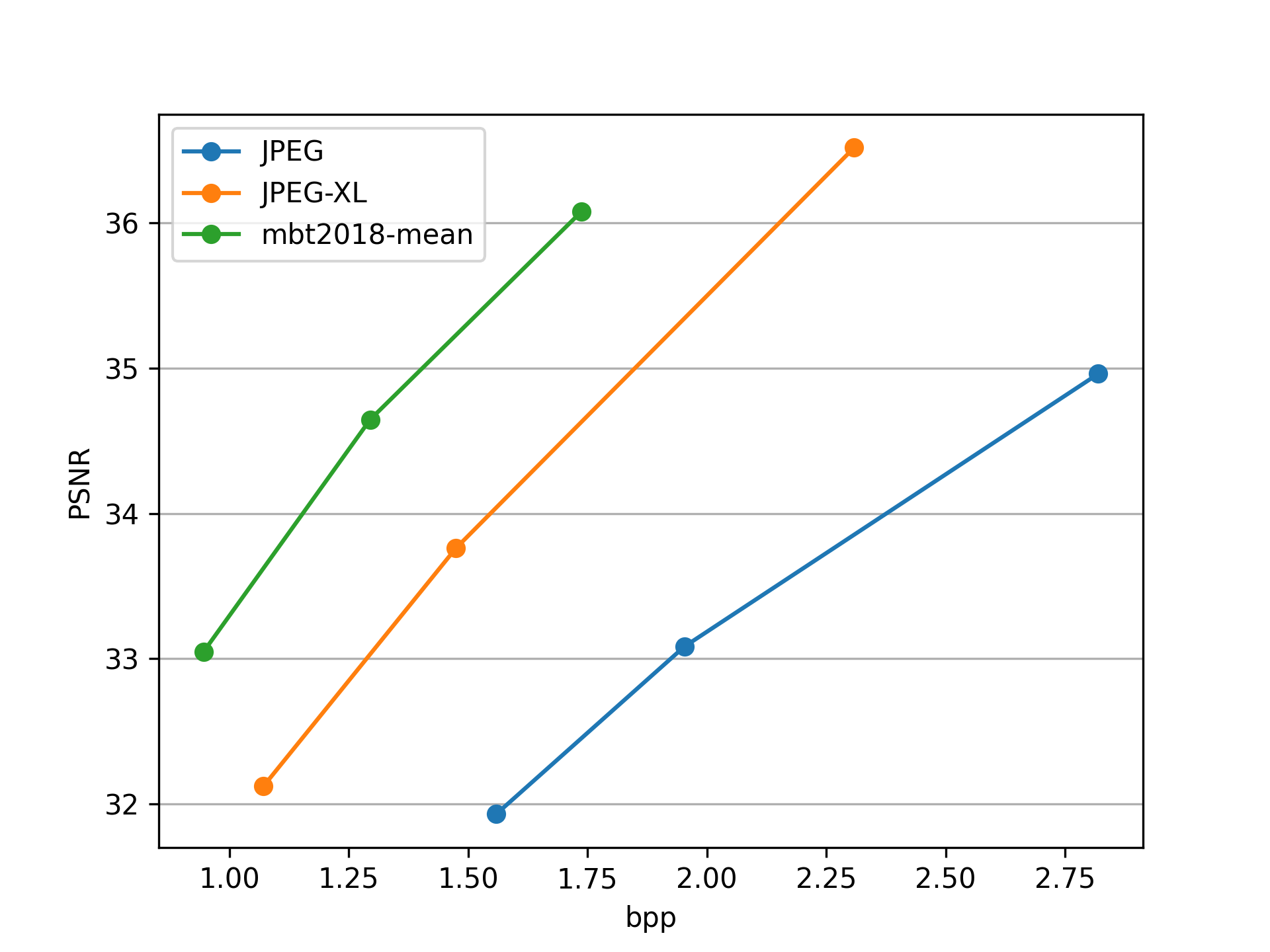}
    \caption{\hl{Rate distortion curve for JPEG, JPEG-XL and the pretrained model mbt2018-mean q7 on the balanced patch dataset. PSNR against bpp. The points for JPEG and JPEG-XL are at quality factors 70, 80, and 90, whereas the deep learning-based model is at quality level 5, 6, and 7. Abbreviations: PSNR = peak signal-to-noise ratio, bpp = bits per pixel.}}
    \label{fig:supmat_bppPSNR}
\end{figure*}

\hl{The rate-distortion curves on the balanced patch test set of HE-stained tissue patches are shown in Fig.} \ref{fig:supmat_bppSSIM}-\ref{fig:supmat_bppPSNR}.

\section{Evaluation on non-HE slide}
\begin{figure*}[h!]
    \centering
    \includegraphics[width=1\linewidth]{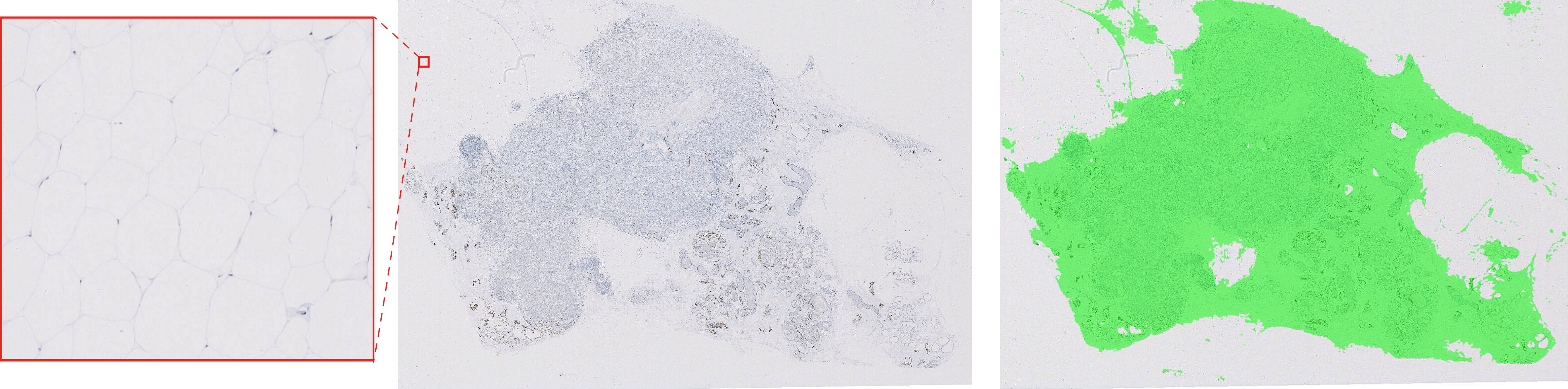}
    \caption{\hl{Original whole slide image (WSI) of a cytokeratin 5/6 stained section from a breast cancer sample (middle) and the tissue segmentation performed with the proposed model at magnification ×2.5 with 10\% overlap (right). The model segmented large parts of the tissue, but missed a considerable portion of the adipose regions. An area from the adipose region is shown in the left figure to illustrate the difficulties with adipose tissue. The WSI in this illustration is the CK56-stained-Case12-IDC from the HE-vs-MPM dataset}~\cite{han2024improving} \hl{available at The Cancer Imaging Archive (TCIA)}~\cite{clark2013cancer}\hl{, with a CC BY 4.0 license, and has been modified for this illustration.}}
    \label{fig:supmat_ck}
\end{figure*}

\hl{The CK 5/6-stained slide CK56-stained-Case12-IDC from the HE-vs-MPM dataset}~\cite{han2024improving} \hl{available at The Cancer Imaging Archive (TCIA)}~\cite{clark2013cancer} \hl{was evaluated to assess the segmentation and compression pipeline on a non-HE stained slide. 

The segmentation model was evaluated visually on the slide at magnification $\times$2.5 with 10\% overlap, Fig.}\ref{fig:supmat_ck}. \hl{The segmentation model performed satisfactory on stromal and epithelial regions, but incorrectly excluded adipose rich areas, Fig. }\ref{fig:supmat_ck}. \hl{ 

The slide was saved with JPEG and JPEG-XL-compression with quality factor 90 using the structure from the original SVS file, creating two new image pyramids with glass. In addition, a corresponding \textit{patch pyramid} of the whole slide including glass was created with patch size 256$\times$256 and saved as PNGs. The patches in the \textit{patch pyramid} were then compressed with the pretrained mbt2018-mean q7 model available in CompressAI, and saved as binary files on disk. The combined size of the binary files was compared with the size of the new JPEG-compressed pyramid. The deep learning-based model saved 56\% space, compared to the corresponding JPEG-compressed image pyramid, whereas JPEG-XL-compression lead to a 22\% size reduction.

Patches of size 256$\times$256 were extracted from within the tissue area, using the proposed tissue segmentation model at $\times$2.5. A balanced set of 112 patches each from magnification $\times$40, $\times$20, $\times$10, $\times$5, $\times$2.5 and $\times$1.25 was created and saved on disk. This number was chosen as the number of patches at magnification $\times$1.25 was 112. The patches were then compressed and decompressed with the pretrained mbt2018-mean q7 model available in CompressAI. The SSIM was calculated between the decompressed patches and the PNG patches. The average SSIM was 0.96, with a min and max of 0.95 and 0.98, respectively. 
}

\section{Recompression runtime}
\hl{The recompression runtimes for JPEG and JPEG-XL (with default settings (compression effort = 7)) on the 21 WSIs, and for the pretrained mbt2018-mean q7 deep learning-based model on the 21 \textit{patch pyramids} are found in Table} \ref{tab:supmat_runtimes}. \hl{The runtimes were measured on the new image pyramids with glass (TIFFs with JPEG-compression, not the original SVS-files) to separate tile reading and decompression times. The recompression runtime is divided into \textit{Read tiles (I/O)} which corresponds to the time to read an original SVS image from disk, \textit{Decompress tiles} which corresponds to the time to decompress the WSI, \textit{Segmentation} which is the time to segment the WSI at $\times$2.5 with 10\% overlap, \textit{Compress} which corresponds to the time to recompress the WSI with the given method, and the \textit{Write (I/O)} which corresponds to the time to write the WSI to disk. All times are reported as the average time for a WSI or \textit{patch pyramid}, including the minimum and maximum. The runtime experiments were performed on a Windows system with the following specifications: Intel Core Processor i7 central processing unit (CPU) with 10 cores, 16GB of RAM, NVIDIA GeForce RTX 4060 (8GB) dedicated GPU, and a solid-state drive (SSD) (NVMe). Segmentation inference utilized TensorRT v8.6. Running the experiments on a hard disk drive (HDD) instead of on an SSD would likely increase the I/O runtimes.}

\section{Additional information}
\begin{equation}
\frac{1}{N}\sum_{n=1}^{N}\left[\left(1-\frac{sum\, bytestrings\, patch_{n}}{JPEG\, compressed\, patch_{n}}\right) * 100\right]
\label{eq:savedspace}
\end{equation}

\begin{table*}[h!]
    \centering
    \begin{adjustbox}{width=\textwidth}
    \begin{tabular}{c|c|c|c|c|c|c}
    \hline
         Organ & Collection & Slide & Slide link & Magnification & Size (px) & License \\
         \hline
         Breast & HER2 tumor ROIs & Her2Neg\_Case\_03 & \href{https://pathdb.cancerimagingarchive.net/caMicroscope/apps/mini/viewer.html?mode=pathdb&slideId=221271}{slide link} & 20x & 23525x29512 & \href{https://creativecommons.org/licenses/by/4.0/}{CC BY 4.0}\\

         Breast& HER2 tumor ROIs & Her2Neg\_Case\_22 & 	\href{https://pathdb.cancerimagingarchive.net/caMicroscope/apps/mini/viewer.html?mode=pathdb&slideId=221290}{slide link} & 20x & 40916x57120 & \href{https://creativecommons.org/licenses/by/4.0/}{CC BY 4.0}\\

         Breast & HER2 tumor ROIs & Her2Pos\_Case\_37 & \href{https://pathdb.cancerimagingarchive.net/caMicroscope/apps/mini/viewer.html?mode=pathdb&slideId=221403}{slide link} & 20x & 59024x40670 & \href{https://creativecommons.org/licenses/by/4.0/}{CC BY 4.0}\\

         Colon & CPTAC-COAD & 05CO004-9fb0c084-16fd-4d54-ab60-d94303 & \href{https://pathdb.cancerimagingarchive.net/caMicroscope/apps/mini/viewer.html?mode=pathdb&slideId=213647}{slide link} & 40x & 63360x60867 & \href{https://creativecommons.org/licenses/by/3.0/}{CC BY 3.0}\\

         Colon & CPTAC-COAD & 05CO060-bdf3565b-acac-4a6a-b2f1-d838be & \href{https://pathdb.cancerimagingarchive.net/caMicroscope/apps/mini/viewer.html?mode=pathdb&slideId=213732}{slide link} & 40x & 38400x45044 & \href{https://creativecommons.org/licenses/by/3.0/}{CC BY 3.0}\\

         Colon & CMB-CRC & MSB-01771-01-02 & 	\href{https://pathdb.cancerimagingarchive.net/caMicroscope/apps/mini/viewer.html?mode=pathdb&slideId=224891}{slide link} & 40x & 45816x61633 & \href{https://creativecommons.org/licenses/by/4.0/}{CC BY 4.0}\\

         Lung & CMB-LCA & MSB-00556-04-06 & 	\href{https://pathdb.cancerimagingarchive.net/caMicroscope/apps/mini/viewer.html?mode=pathdb&slideId=225029}{slide link} & 40x & 67727x60494 & \href{https://creativecommons.org/licenses/by/4.0/}{CC BY 4.0}\\

         Lung & CMB-LCA & MSB-02118-01-07 & \href{https://pathdb.cancerimagingarchive.net/caMicroscope/apps/mini/viewer.html?mode=pathdb&slideId=224896}{slide link} & 40x & 89639x77391 & \href{https://creativecommons.org/licenses/by/4.0/}{CC BY 4.0}\\

         Lung & CMB-LCA & MSB-02151-02-13 & \href{https://pathdb.cancerimagingarchive.net/caMicroscope/apps/mini/viewer.html?mode=pathdb&slideId=311873}{slide link} & 40x & 85656x72466 & \href{https://creativecommons.org/licenses/by/4.0/}{CC BY 4.0}\\

         Ovary & CPTAC-OV & 20OV005-65e6e2d2-5d7d-4ef1-898c-149b90 & \href{https://pathdb.cancerimagingarchive.net/caMicroscope/apps/mini/viewer.html?mode=pathdb&slideId=217306}{slide link} & 40x & 72960x45107 & \href{https://creativecommons.org/licenses/by/3.0/}{CC BY 3.0}\\

         Ovary & CPTAC-OV & 22OV001-82eb0c9b-fb7f-436f-b196-be0003 & \href{https://pathdb.cancerimagingarchive.net/caMicroscope/apps/mini/viewer.html?mode=pathdb&slideId=217308}{slide link} & 40x & 32640x34688 & \href{https://creativecommons.org/licenses/by/3.0/}{CC BY 3.0}\\

         Ovary & CPTAC-OV & 02OV005-c5c362f6-f42a-49c2-a16d-3ec5ab & \href{https://pathdb.cancerimagingarchive.net/caMicroscope/apps/mini/viewer.html?mode=pathdb&slideId=217146}{slide link} & 40x & 31680x28862 & \href{https://creativecommons.org/licenses/by/3.0/}{CC BY 3.0} \\

         Pancreas & CPTAC-PDA & C3N-00513-23 & 	\href{https://pathdb.cancerimagingarchive.net/caMicroscope/apps/mini/viewer.html?mode=pathdb&slideId=217665}{slide link} & 20x	& 11951x9917 & \href{https://creativecommons.org/licenses/by/3.0/}{CC BY 3.0}\\

         Pancreas & CPTAC-PDA & C3L-02897-21 & 	\href{https://pathdb.cancerimagingarchive.net/caMicroscope/apps/mini/viewer.html?mode=pathdb&slideId=217531}{slide link} & 20x	& 23903x31106 & \href{https://creativecommons.org/licenses/by/3.0/}{CC BY 3.0}\\

        Pancreas & CPTAC-PDA & C3L-00395-23 & \href{https://pathdb.cancerimagingarchive.net/caMicroscope/apps/mini/viewer.html?mode=pathdb&slideId=217344}{slide link} & 20x & 21911x17218 & \href{https://creativecommons.org/licenses/by/3.0/}{CC BY 3.0}\\

         Prostate & CMB-PCA & MSB-06184-07-02 & \href{https://pathdb.cancerimagingarchive.net/caMicroscope/apps/mini/viewer.html?mode=pathdb&slideId=224972}{slide link} & 40x & 29879x50979 & \href{https://creativecommons.org/licenses/by/4.0/}{CC BY 4.0}\\

         Prostate & CMB-PCA & MSB-08965-01-02 & \href{https://pathdb.cancerimagingarchive.net/caMicroscope/apps/mini/viewer.html?mode=pathdb&slideId=312082}{slide link} & 40x	& 73703x26724 & \href{https://creativecommons.org/licenses/by/4.0/}{CC BY 4.0}\\

         Prostate & CMB-PCA & MSB-01956-01-20 & 	\href{https://pathdb.cancerimagingarchive.net/caMicroscope/apps/mini/viewer.html?mode=pathdb&slideId=312479}{slide link} & 40x	& 23903x9335 & \href{https://creativecommons.org/licenses/by/4.0/}{CC BY 4.0}\\

         Uterus & CPTAC-UCEC & C3L-00767-21 & 	\href{https://pathdb.cancerimagingarchive.net/caMicroscope/apps/mini/viewer.html?mode=pathdb&slideId=218260}{slide link} & 	20x	& 41831x35880 & \href{https://creativecommons.org/licenses/by/3.0/}{CC BY 3.0}\\

         Uterus	& CPTAC-UCEC & C3L-00781-27	& \href{https://pathdb.cancerimagingarchive.net/caMicroscope/apps/mini/viewer.html?mode=pathdb&slideId=218281}{slide link} & 20x	& 17927x21295 & \href{https://creativecommons.org/licenses/by/3.0/}{CC BY 3.0}\\

         Uterus & CPTAC-UCEC & C3L-01282-26 & 	\href{https://pathdb.cancerimagingarchive.net/caMicroscope/apps/mini/viewer.html?mode=pathdb&slideId=218344}{slide link} & 20x	& 33863x20290 & \href{https://creativecommons.org/licenses/by/3.0/}{CC BY 3.0}\\
         \hline
         \hl{Breast} & \hl{HE-vs-MPM} & \hl{CK56-stained-Case12-IDC} & \href{https://pathdb.cancerimagingarchive.net/caMicroscope/apps/viewer/viewer.html?slideId=226278&mode=pathdb}{\hl{slide link}} & \hl{40x} & \hl{163680x100416} & \href{https://creativecommons.org/licenses/by/4.0/}{CC BY 4.0}\\
        \hline
    \end{tabular}
    \end{adjustbox}
    \caption{Dataset used for glass removal and training of tissue compression models~\cite{Farahmand22,CPTAC_coad_20,CMB-CRC22,CMB-LCA22,CPTAC-OV20,CPTAC-PDA18,CMB-PCA22,CPTAC-UCEC19,han2024improving}.}
    \label{tab:dataset}
\end{table*}

\begin{table}[h!]
    \centering
    \begin{tabular}{c|c}
    \hline
         Magnification & Number of patches \\
         \hline
         $\times$1.25& 190  \\
         $\times$2.5& 916  \\
         $\times$5& 3 995  \\
         $\times$10& 16 524  \\
         $\times$20& 17 078 \\
         $\times$40& 16 515 \\
        \hline
        
    \end{tabular}
    \caption{Number of patches included in the tissue segmentation dataset at different magnifications.}
    \label{tab:patchnumber}
\end{table}

\begin{table}[h!]
    \centering
    \begin{tabular}{c c c c c c}
    \hline
         Set & Organ & Slide & Image level & Patch size & Patches (N)\\
         \hline
         Train & Lung & MSB-00556-04-06.svs & 0 & 256 & 21 392\\
         Train & Colon & MSB-01771-01-02.svs & 3 & 256 & 12\\
         Train & Breast & Her2Pos\_Case\_37.svs & 1 & 256 & 1 140\\
         Train & Lung & MSB-02118-01-07.svs & 2 & 256 & 222\\
         Train & Ovary & 22OV001\_82eb0c9b-fb7f-436f-b196-be0003.svs & 2 & 256 & 28\\
         Train & Lung & MSB-02151-02-13.svs & 0 & 256 & 7 350\\
         Train & Pancreas & C3N-00513-23.svs & 0 & 256 & 387\\
         Train & Prostate & MSB-01956-01-20.svs & 1 & 256 & 46\\
         Train & Prostate & MSB-08965-01-02.svs & 0 & 256 & 4 146\\
         Train & Uterus & C3L-00781-27.svs & 2 & 256 & 17\\
         Train & Ovary & 20OV005\_65e6e2d2-5d7d-4ef1-898c-149b90.svs & 2 & 256 & 77\\
         Train & Prostate & MSB-06184-07-02.svs & 0 & 256 & 2 060\\
         Train & Colon & 05CO060-bdf3565b-acac-4a6a-b2f1-d838be.svs & 0 & 256 & 14 073\\
         Train & Pancreas & C3L-00395-23.svs & 0 & 256 & 1 344\\
         Val & Breast & Her2Neg\_Case\_03.svs & 0 & 256 & 1 676\\
         Val & Pancreas & C3L-02897-21.svs & 2 & 256 & 11\\
         Val & Colon & 05CO004-9fb0c084-16fd-4d54-ab60-d94303.svs & 3 & 256 & 20\\
         Test & Ovary & 02OV005\_c5c362f6-f42a-49c2-a16d-3ec5ab.svs &  & 256 & \\
         Test & Breast & Her2Neg\_Case\_22.svs &  & 256 & \\
         Test & Uterus & C3L-00767-21.svs &  & 256 & \\
         Test & Uterus & C3L-01282-26.svs &  & 256 & \\
         \hline
    \end{tabular}
    \caption{Data split of HE-stained histology data into train, validation and test sets used for training the tissue compression models ~\cite{Farahmand22,CPTAC_coad_20,CMB-CRC22,CMB-LCA22,CPTAC-OV20,CPTAC-PDA18,CMB-PCA22,CPTAC-UCEC19}.}
    \label{tab:compression_dataset}
\end{table}

\begin{table*}[h!]
    \centering
    \begin{tabular}{c|c}
        \hline
        Magnification & Number of patches \\
        \hline
         $\times$1.25 & 138\\
         $\times$2.5 & 565\\
         $\times$5 & 2 296\\
         $\times$10 & 9 168 \\
         $\times$20 & 36 947\\
         $\times$40 & 6 045\\
         $\times1.25$, $\times2.5$, $\times5$, $\times10$, $\times20$, $\times40$ & 828\\
         \hline
    \end{tabular}
    \caption{The number of patches generated at the different magnifications used in the test set (4 slides) for validating the tissue compression models. The last row corresponds to the balanced test set with an equal number of patches from each magnification. Only WSIs which were scanned at $\times40$ were used to generate patches from magnification $\times40$. The patches were generated as PNG, JPEG 90, JPEG-XL 90, and JPEG-2000.}
    \label{tab:test_data}
\end{table*}

\begin{table*}[h!]
    \centering
    \begin{adjustbox}{width=\textwidth}
    \begin{tabular}{l c c l c c c c c c}
        \toprule
         \hl{Read tiles (I/O) [s]} & \hl{Decompress tiles [s]} & \hl{Segmentation [s]} & \hl{Method }   & \hl{Compress [s]} & \hl{Write (I/O) [s]} & \hl{Other [s]} & \hl{Total [s]}\\
         \midrule
                                     &                                   &        \hl{NA}           & \hl{JPEG Glass}                  &  \hl{4 (0.2, 16)}    &    \hl{7 (0.1, 34)}   &  \hl{6 (0.3, 42)}        &  \hl{26 (1, 134)}     \\
            \hl{5 (0.1, 35)}         &    \hl{5 (0.2, 20)}               &   \hl{0.9 (0.3, 2)}      & \hl{JPEG single-color}      &  \hl{4 (0.2, 15)}    &    \hl{6 (0.1, 35)}   &  \hl{19 (1, 74)}       &  \hl{39 (3, 170)}     \\
                                     &                                   &   \hl{0.9 (0.3, 2)}      & \hl{JPEG Empty tiles}       &  \hl{2 (0.1, 11)}    &    \hl{3  (0.05, 46)} &  \hl{22 (1, 112)}       &  \hl{39 (2, 226)}      \\
        \midrule    
                                     &                                   &       \hl{NA}           & \hl{JPEG-XL Glass}               &  \hl{788 (49, 2598)} &    \hl{9 (0.2, 52)}   &  \hl{9 (0.6, 43)}       &  \hl{835 (51, 2786)}    \\
            \hl{21 (0.7, 91)$^*$}    &    \hl{6 (0.3, 23)}               &   \hl{0.9 (0.3, 2)}      & \hl{JPEG-XL single-color}   &  \hl{857 (50, 2796)} &    \hl{9 (0.2, 51)}  &  \hl{26 (2, 97)}       &  \hl{926 (54, 3059)}    \\
                                     &                                   &   \hl{0.9 (0.2, 2)}      & \hl{JPEG-XL Empty tiles}    &  \hl{298 (18, 1447)} &    \hl{3  (0.07, 28)} &  \hl{23 (2, 90)}       &  \hl{344 (21, 1647)}    \\
        \midrule  
            \multicolumn{2}{c}{\hl{126 (5, 495)$^{**}$}}                 &                   & \hl{DL Glass}                    &   \hl{558 (31, 1858)}      &    \hl{33 (3, 93)$^\diamond$}   &  \hl{4 (0.2, 14)}      &  \hl{721 (39, 2450)}       \\
            \multicolumn{2}{c}{\hl{108 (4, 450)$^{**}$}}                 &                   & \hl{DL single-color}             &   \hl{561 (31, 1878)}      &    \hl{33 (3, 91)$^\diamond$}     &  \hl{3 (0.1, 9)}      &  \hl{704 (38, 2422)}        \\
            \multicolumn{2}{c}{\hl{57 (2, 308)$^{**}$}}                  &                   & \hl{DL Empty tiles}              &   \hl{212 (11, 1053)}      &   \hl{13 (0.7, 46)$^\diamond$}     & \hl{1 (0.1, 5)}        &  \hl{284 (15, 1411)}       \\
         \bottomrule
    \end{tabular}
    \end{adjustbox}
    \caption{\hl{Recompression runtime for JPEG, JPEG-XL, and the pretrained mbt2018-mean q7 deep learning model. The runtimes are the average runtimes per WSI for JPEG and JPEG-XL and per \textit{patch pyramid} for the deep learning model. The minimum and maximum values are reported in parentheses. The empty tiles in the empty \textit{patch pyramids} were removed from the patch set. The segmentation time is not included for images with glass as no segmentation was performed when glass was not removed. *) An increased read tile runtime was observed when using JPEG-XL. The reason for this is unknown, but may be due increased memory usage of JPEG-XL influencing caching of the TIFF pyramids. **) \textit{Read tiles (I/O)} and \textit{Decompress tiles} are not reported as separate values for the deep learning model as deep learning-based compression was performed on patch pyramids with PNG patches, and read with the image library Pillow (PIL). Segmentation time was not reported for the deep learning-based model as glass areas were already replaced by single-colored pixels or glass tiles were already removed from the patch pyramids. $\diamond$) DL compressed patches were stored as two binary files per patch. The runtime experiments were performed on a Windows system with the following specifications: Intel Core Processor i7 central processing unit (CPU) with 10 cores, 16GB of RAM, NVIDIA GeForce RTX 4060 (8GB) dedicated GPU, and a solid-state drive (SSD) (NVMe). Segmentation inference utilized TensorRT v8.6.}}
    \label{tab:supmat_runtimes}
\end{table*}

\clearpage
\printbibliography